\documentclass[lettersize,journal]{IEEEtran}
\usepackage{amsmath,amsfonts}
\usepackage{algorithmic}
\usepackage{algorithm}
\usepackage{array}
\usepackage[caption=false,font=normalsize,labelfont=sf,textfont=sf]{subfig}
\usepackage{textcomp}
\usepackage{booktabs}
\usepackage{stfloats}
\usepackage{url}
\usepackage{verbatim}
\usepackage{graphicx}
\usepackage{cite}
\usepackage{multirow}
\usepackage{booktabs}
\begin{document}

\title{HSONet:A Siamese foreground association-driven hard case sample optimization network for high-resolution remote sensing image change detection}

\author{Chao Tao, Dongsheng Kuang, Zhenyang Huang, Chengli Peng, Haifeng Li,~\IEEEmembership{Member,~IEEE,}
\thanks{The work was supported in part by the Major Program Project of Xiangjiang Laboratory under Grant 22XJ01010, in part by the National Natural Science Foundation of China under Grant 61973047 and Grant 42171458, and in part by using Computing Resources at the High-Performance Computing Platform of Central South University. (Corresponding author: Haifeng Li.)}
\thanks{Chao Tao, Dongsheng Kuang, Zhenyang Huang, Chengli Peng, Haifeng Li are with the School of Geosciences and Info-Physics, Central South University, Changsha 410083, China, and also with the Xiangjiang Laboratory, Changsha 410205, China.}
}

\markboth{Journal of \LaTeX\ Class Files,~Vol.~14, No.~8, August~2021}%
{Shell \MakeLowercase{\textit{et al.}}: A Sample Article Using IEEEtran.cls for IEEE Journals}


\maketitle

\begin{abstract}
Deep learning technologies have driven significant advances in change detection (CD) techniques. RS-CD relies on the model's ability to learn features of marked change objects, known as foreground targets. In addition to foreground targets, the most valuable samples in the image space for model optimization are the unlabelled and semantically ambiguous samples in the background, which include hard samples, pseudochanges, and noninteresting changes; these samples are collectively referred to as "hard case samples" in this paper. In the later training stages, further improvement of the model's ability to determine changes relies on how well the model learns hard cases; however, there are two additional challenges to learning hard case samples: (1) change labels are limited and tend to pointer only to foreground targets, yet hard case samples are prevalent in the background, which leads to optimizing the loss function focusing on the foreground targets and ignoring the background hard cases, which we call 'imbalance'. (2) Complex situations, such as light shadows, target occlusion, and seasonal changes, induce hard case samples, and in the absence of both supervisory and scene information, it is difficult for the model to learn hard case samples directly to accurately obtain the feature representations of the change information, which we call 'missingness'. We propose a Siamese foreground association-driven hard case sample optimization network (HSONet). To deal with this imbalance, we propose an equilibrium optimization loss function to regulate the optimization focus of the foreground and background, determine the hard case samples through the distribution of the loss values, and introduce dynamic weights in the loss term to gradually shift the optimization focus of the loss from the foreground to the background hard cases as the training progresses. To address this missingness, we understand hard case samples with the help of the scene context, propose the scene-foreground association module, use potential remote sensing spatial scene information to model the association between the target of interest in the foreground and the related context to obtain scene embedding, and apply this information to the feature reinforcement of hard cases. Experiments on four public datasets show that HSONet outperforms current state-of-the-art CD methods, particularly in detecting hard case samples.
\end{abstract}


\section{Introduction}
\IEEEPARstart{C}{hange} detection is a crucial technology in interpreting and analyzing remote sensing images. Remote sensing change detection (RS-CD) focuses on the same geographical areas on the Earth's surface, to capture changes in land cover information over time and to generate binary change maps representing differential information. Today, a multitude of CD algorithms and theoretical models have been proposed. Coupled with the development of remote sensing imaging technology, RS-CD has been widely applied in many significant fields, including land resource management\cite{1}, natural disaster monitoring\cite{2}, agricultural land analysis\cite{3}, and urban-rural development planning\cite{4,5}.

In recent years, owing to the gradual maturation of the commercialization of high-resolution remote sensing satellites, current high-resolution remote sensing imagery (HRRS) can describe a variety of feature entities more comprehensively and in more detail\cite{6,7}. However, the prevalence of light shading, target occlusion, and seasonal changes in optical remote sensing images, coupled with difficulties such as spectral feature confusion, increases the intraclass variance in features, while the interclass variance decreases, which in turn leads to a significant reduction in the separability of remotely sensed targets. This problem is particularly pronounced in HRRS and, therefore, induces intractable CD problems involving hard samples, pseudochanges, and noninteresting changes. Like in semantic segmentation tasks, RS-CD relies heavily on the effectiveness of feature learning for key objects, which are also called foreground targets. In addition to these foreground targets, the most valuable elements for model optimization in the image space are the semantically ambiguous samples in the background, i.e., hard case samples, which include hard samples, pseudochanges, and noninteresting changes. After many experimental explorations, we realized that hard case samples are often more valuable for optimization than foreground targets are in the later model training stages because, for the pixel-level prediction task, the deep neural network first obtains the foreground target from the background information and then calculates the conditional probability of the category to which the foreground target belongs on a pixel-by-pixel basis and selects the highest probability score as the category information for that pixel. The probability of a hard case sample being correctly discriminated can increase if and when the hard case target and the foreground target are optimized to the same degree. In contrast, hard case samples interfering with foreground target feature extraction and feature discrimination processes will have a negative impact on RS-CD, so optimizing the feature learning process for background hard cases will certainly be highly important.

At present, deep learning techniques are increasingly widely used in remote sensing, and deep neural network-based change detection models are also focused on solving hard case sample problems and subproblems. These methods can be broadly categorized into two groups: (1) strengthening the feature representations of key information and (2) adding an attention mechanism and its variants. In terms of enhancing feature representation, Zheng et al.\cite{8} considered the importance of the accuracy and completeness of building boundary recognition for CD and proposed a target edge guidance module and a feature differential enhancement module based on a transformer for refining edge features and fusing different levels of change information, respectively, thus enhancing the high-frequency information of buildings. Xu et al.\cite{9} realized the impact of the structural information of remote sensing targets on the detection accuracy and proposed a multiscale context aggregation network in which the global attention pyramid module and the dense feature fusion module are used to enhance the depth features of the original target and bridge the semantic gap between multiscale features, respectively. In terms of incorporating attention mechanisms, Fang et al.\cite{10} introduced an integrated channel attention module for the deep supervision of interest features and designed a tightly coupled information transmission mechanism between encoders and decoders. This approach mitigates the loss of deep localization information in neural networks and ultimately enables the localization of edge pixels and the capture of small targets. Chen et al.\cite{11} considered the prevalent pseudochange problem in CD and solved the pseudochange problem to a certain extent by applying spatial and channel attention mechanisms to sensitize the network to changes in interest and additionally balancing the variability between samples by penalizing noninteresting changes and increasing the attention given to changes in interest. In the first category of methods, Chen et al.\cite{12} used differential features at different scales to strengthen the feature representation of changes, effectively mitigating issues such as pseudochanges. However, the greatest challenge in optical remote sensing image feature extraction lies in learning accurate semantic features from hard case samples, and strengthening only the feature representation and designing some kind of attentional mechanism cannot completely guarantee the accuracy and completeness of the semantic understanding of hard case samples; thus, so there is currently no learning methods available for hard case samples. However, networks such as DMINet\cite{13} and DASNet\cite{11} have addressed several challenges, such as pseudochanges and differences between foreground and background samples, and have optimized the focus of the network using joint attention and spatial and channel attention mechanisms, thus gaining some robustness against pseudochange issues. However, changing the network's attention to various types of samples does not improve the model's ability to identify hard case samples because the hard case samples tend to be weak, dispersed, or small in volume in the feature space; moreover, it is not easy for them to dominate the gradient optimization during the training process. Additionally, they will not be subjected to the same optimization effort as the foreground samples, and they will not be easy to focus on the attention mechanism.

In the HRRS, hard case samples in the background are often represented in the image space as light shadows, object occlusions, small targets, etc. We realize that each of these cases can be considered as some form of 'camouflage' of a simple sample, e.g., 'occlusion' is the covering of a part of the target with a mask, e.g., the tree in the red box in Figure 1 occludes the main body of the road. 'Shadows' are dark areas on the same side of a 3-D target, such as the shaded area around a high-rise building in the yellow box in Figure 1, while 'small targets' can be regarded as scaled-down versions of conventional targets, such as the discrete distribution of remote sensing targets in the orange box in Figure 1. Based on the above thinking, learning hard case samples by the network can start from 'simple samples', and whenever the model obtains an excellent learning effect on simple samples, it gradually begins learning background hard cases. Experiments have proven that this process optimization will obtain a better learning effect on hard case samples. This is in line with the core idea of curriculum learning\cite{14,15,pengjian}, which allows the model to learn relatively simple samples at first so that it can form basic concepts and patterns and then gradually present them with more challenging hard samples as the model continues to mature. A large body of work has verified that this strategy is beneficial for learning complex concepts via neural networks.
\begin{figure}[H]
    \includegraphics[width=0.5\textwidth]{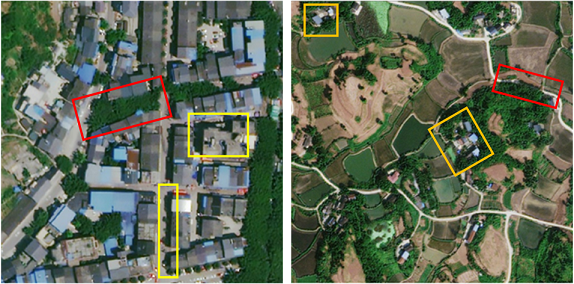}
    \caption{Sample display of hard case sample in remote sensing image. Where hard case samples such as target occlusion, shadows and small targets are shown in red, yellow and orange boxes respectively.}
    \label{fig1}
\end{figure}
After analysing the aforementioned issues, we propose a Siamese foreground association-driven hard case sample optimization network for remote sensing image change detection. Using scene context to deepen the understanding of background hard cases is an excellent strategy; therefore, we propose the foreground-scene association module, which uses potential remote sensing spatial scene information to model the relationship between interest targets in the foreground and the associated context and activate the output of hard case samples in the background to increase the difference between the foreground and the background and, conversely, to reduce the difference between the foreground and the hard cases of interest and to solve leakage and misreporting in CD. In contrast, background samples are less likely to dominate the gradient during training, but hard case samples are more valuable for model optimization in the later stages of training; however, the number of background hard cases is much smaller than the number of simple samples. To address this problem, we propose an equilibrium optimization loss function(EO-loss) so that the network focuses on the foreground target in the pretraining stage and the background hard cases in the posttraining stage for balanced optimization. We conducted relevant experiments on four public change detection datasets, and the results demonstrated the efficiency and robustness of HSONet. The main contributions of this paper are as follows:
\begin{enumerate}
  \item{To address the common issue of hard case samples in HRRS, we propose a Siamese foreground association-driven hard case sample optimization network for remote sensing image change detection. To address the 'imbalance' in optimizing hard case samples, we introduce an equilibrium optimization loss function. By analysing the distribution of loss values, hard case samples are identified, and dynamic weights are introduced in the loss term. This gradually shifts the focus of loss optimization from the foreground to background hard cases, ensuring the successful discovery of interest hard case samples.}
  \item{To better understand the semantics of hard case samples, we propose a foreground-scene association module. It uses latent remote sensing spatial scene information to model the association between foreground targets of interest and related context, obtaining scene embeddings. These are then applied to the feature discrimination of hard case samples. Moreover, this approach reduces the feature difference between interest samples in background hard cases and the foreground, thereby suppressing issues such as missed detections in CD.}
  \item{Compared to 11 advanced baseline methods, HSONet achieves state-of-the-art results on four datasets, CDD\cite{16}, LEVIR-CD\cite{17}, Google-CD\cite{18}, and SYSU-CD\cite{19}, with absolute F1 scores reaching 92.42\%, 98.11\%, 93.14\%, and 82.84\%, respectively. Compared to the latest method, USSFC-Net\cite{20} yields improvements of 2.29\%, 2.61\%, 6.34\%, and 1.77\%, with an average increase of 3.26\%.}
\end{enumerate}

The other sections of this paper are as follows. Section 2 reviews related work. Section 3 describes our method. Section 4 describes the design of the validation experiments, followed by the analysis and discussion. Section 5 concludes the paper and proposes ideas and suggestions for future research.

\section{Related works}
In recent years, deep learning technology has taken remote sensing image change detection to a new level, with a large number of DL-based CD algorithms and theoretical models being successively proposed. In the following text, we review two parts of the related work: CD methods based on two mainstream deep networks (CNNs and transformers) and the current status and issues of optimization methods for hard case samples.

\subsection{CD methods based on CNNs or Transformers}
The introduction of convolutional neural networks (CNNs) has driven rapid developments in RS-CD research, with existing studies divided into two main stages: (1) the optimization stage of the network structure and (2) the application stage of attention mechanisms. In the network structure optimization stage, CNN-based methods have evolved towards depth and modularity, aiming to obtain more abstract feature representations and achieve plug-and-play functionality, which aids in understanding complex objects and scenes. For instance, Zhan et al.\cite{21} first organized the CNN backbone within a Siamese framework to obtain feature information and subsequently determined changes by measuring the distance between features and threshold segmentation. In 2018, Daudt et al.\cite{22} proposed three FCN-based CD methods—FC-EF, FC-Siam-diff, and FC-Siam-conc—and discussed the impact of network structures and feature fusion methods on CD tasks through comparisons. Numerous FCN-based CD networks have been proposed; for example, Guo et al.\cite{23} proposed a fully convolutional metric network that learns implicit metric criteria to measure the similarity of feature mappings, thereby learning discrimination patterns for change and no change; Zheng et al.\cite{24} introduced an end-to-end U-shaped FCN change detection network called CLNet, which combines features of different scales and multilevel context information to enhance CD effectiveness; and Peng et al.\cite{25} proposed a difference-enhanced dense-attention convolutional neural network, which enhances the accuracy of change information retrieval by introducing dense attention and the DE unit. In the application stage of attention mechanisms, numerous works guide the network to learn emphasized part features through attention mechanisms, thus learning key discriminative information related to changes. For example, Chen et al.\cite{11} introduced a dual-attention mechanism (spatial and channel attention) into a fully convolutional Siamese network, solving problems such as pseudochanges by enhancing the focus on interest changes.

To address the complexity and semantic uncertainty of remote sensing targets, recent years have seen a surge of work introducing context modelling to enhance the learning capability of neural networks, which has proven to be crucial for RS-CD tasks. Current context modelling methods include four mainstream methods: multiscale feature fusion, a deep supervision network architecture, the application of dilated convolution, and the design of various attention mechanisms. Many works combine the advantages of mainstream methods, integrating them into a single network. For example, Jiang et al.\cite{26} used various methods to merge low-level and coattention-level features and establish long-range contextual connections. Wang et al.\cite{27} proposed a depth-supervised network based on self-attention that extracts multilevel image features during the encoding phase. To further highlight change features, they proposed an adaptive attention mechanism that combines spatial and channel features, capturing the relationships between change features of different scales. Using CNNs to obtain multiscale feature mappings of targets is significant for CD tasks, but fundamentally, CNNs lack the capability to model long-range dependencies. Fortunately, the advent of the transformer has changed this situation.

Dosovitskiy et al.\cite{28} first applied a transformer to image classification tasks in 2018, achieving remarkable results; moreover, the transformer also applied RS-CD to new heights. Chen et al.\cite{29} were the first to apply transformers to CD tasks and proposed the bitemporal image transformer (BIT) network. Its encoder, embedded with a transformer, models the spatiotemporal context of abstract feature mappings obtained from CNNs. Then, tokens enriched with contextual information are fed back to the pixel space through the transformer decoder to locate changes in interest. BIT once again proved the superiority of the transformer in context modelling. Since contextual information is crucial for semantically understanding remote sensing imagery, many RS-CD networks based on ViT have been proposed. Currently, transformer-based methods can be divided into two main categories: hybrid methods of transformers and CNNs. For example, Feng et al.\cite{30} reported that the sequential use of a CNN and ViT hindered the interaction of depth and breadth features; consequently, they proposed a parallel method to realize the coupling and complementarity of local and global features. Li et al.\cite{31} found the advantages of U-Net and ViT to be complementary, creatively embedding ViT into a U-shaped network, overcoming the difficulty of feature layer relationship modelling and the inaccuracy of differential feature representation. Xu et al.\cite{32} adopted progressive sampling ViT to eliminate irrelevant change interference to address pseudochanges and missed detections and subsequently applied a fusion module to obtain complete edge information, ensuring the accuracy of the change information. The other category includes pure transformer-based CD methods, such as those used by Yan et al.\cite{33}, who introduced a pyramid structure to aggregate multilevel features of the transformer for solving irregular change area boundaries using a progressive attention module to enhance the representation level of interdependent features. Similarly, Zhang et al.\cite{34} designed a dual U-shaped CD network using the Swin-T backbone, aiming to overcome the inherent limitations of convolutional operations completely and fully exploit the global modelling advantage of the transformer.

\subsection{Current research status of optimization methods for hard case samples}
Hard case sample problems are widespread in machine learning and data science, typically referring to situations where certain samples are more difficult to classify, learn, and predict correctly during model training. Several aspects of the hard case sample problem in remote sensing imagery can be summarized as follows: 1) sample imbalance, 2) small targets, 3) shadow coverage and object occlusion, and 4) pseudochanges. Many studies have explored the above problems in computer vision tasks; for example, for the sample imbalance problem, the current mainstream approach is to optimize loss function design. Lin et al.\cite{35} proposed the focal loss for object detection tasks, introducing an adjustable focus parameter to reduce the weight of easy-to-classify samples, focusing more on hard case samples, while weighted cross-entropy loss\cite{36} adjusts the model's focus on different categories by introducing different weights for each class, thus solving the sample imbalance problem. Zhao et al.\cite{37} introduced Dice loss in medical image segmentation tasks, minimizing the Dice coefficient to increase the sensitivity of the model to situations with fewer pixels and suppress the background weight; however, Dice loss can be used only for binary segmentation tasks. For issues such as small targets and pseudochanges, in addition to data augmentation operations, optimizing the model structure is also crucial. For instance, Lin et al.\cite{38} proposed the FPN method, which generates feature maps of different resolutions at different levels of the network. High-level features contain stronger semantics, and low-level features contain more details, enhancing the perception of small targets. Fu et al.\cite{39} introduced a dual-attention mechanism in semantic segmentation tasks, using spatial attention to capture key spatial information and channel attention to capture features more relevant to the task, thereby improving key information extraction. Of course, research on CD has also actively explored optimization solutions for hard case samples. For example, MFCN\cite{40} designs multiscale convolutional kernels to extract detailed surface features and combines the WBCE and Dice loss for balanced sample optimization, improving the model's ability to discriminate minute features from two aspects. Zhu et al.\cite{41} also focused on problems such as small target changes and edge pixel misclassification, proposing ECFNet and designing three processes—feature extraction, feature comparison, and feature fusion. In the fusion process, the number of channels is constrained to better utilize the fine-grained information in multiscale features for result prediction. Guo et al.\cite{42} realized the importance of interest feature extraction and feature fusion for CD and proposed three modules, deep multiscale feature extraction, parallel convolutional feature fusion, and self-attention-based feature refinement, to integrate multiscale change information and further enhance the feature representation of interest targets. The current optimization methods for hard case samples mainly focus on model structure improvement, interaction of deep and shallow layer information, and optimization of attention mechanisms. However, the idea of this paper is to start from the learning rules of hard case samples and divide the feature learning process according to difficulty characteristics. Sample optimization can be achieved through two steps: hard case sample mining and balanced optimization of foreground and hard cases. In this process, background hard cases are enhanced using relevant scene information in the foreground, thus improving feature discriminability and accuracy. Therefore, we designed the foreground-scene association module and an EO-loss to learn hard case samples commonly found in the background.

\section{Methodology}
In this chapter, we provide a detailed introduction to the proposed CD method. Section A presents the overall framework of HSONet. Section B introduces the variant feature pyramid network encoder. Section C describes the foreground-scene association module. Section D covers the dual-temporal feature fusion and feature decoding structure. Section E introduces the EO-loss.

\begin{figure*}[ht]
\centering
\includegraphics[width=7in]{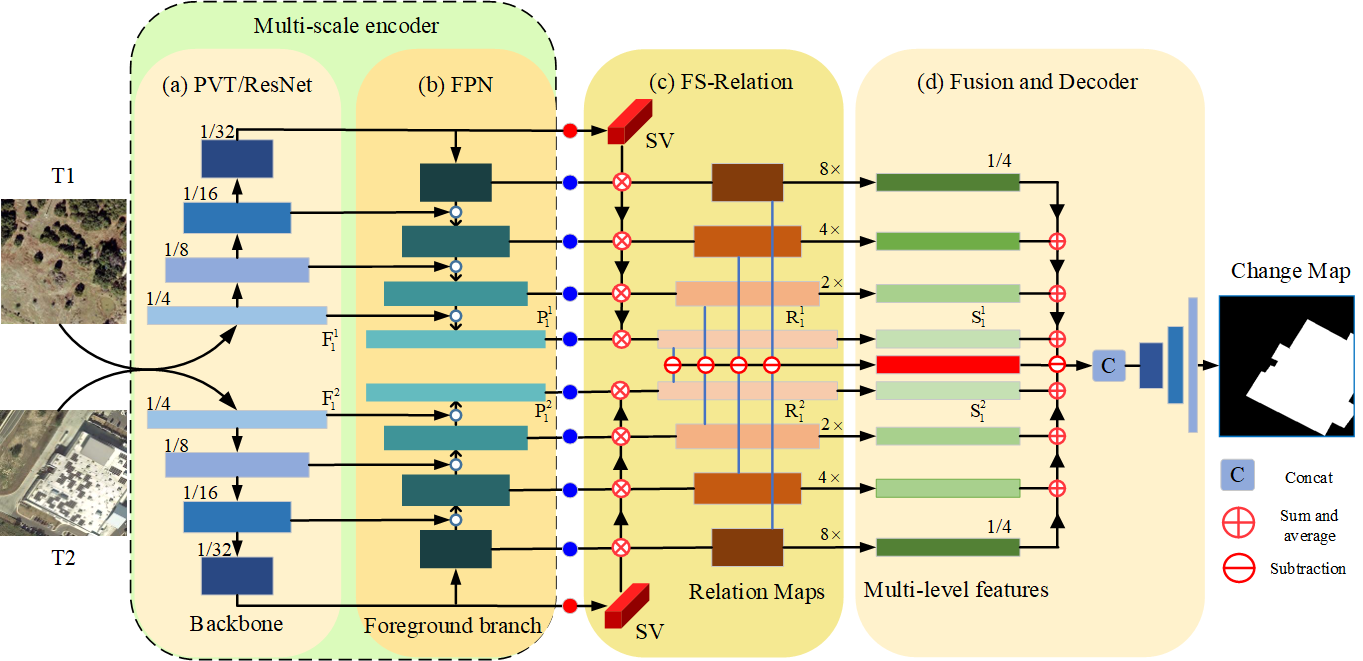}
\caption{Overview of the HSONet.}
\label{fig2}
\end{figure*}

\subsection{Network Architecture}
As shown in Figure 2, the overall structure of HSONet is an end-to-end Siamese network. A pair of dual-temporal remote sensing images is taken as input, and a binary change map is output. The network consists of a variant feature pyramid network module, a foreground-scene association module (FS-relation), an EO-loss, and a feature fusion and lightweight decoder. The V-FPN module is responsible for multiscale target feature extraction and obtaining deep feature information. The FS-relation module is built on a collection of multiscale feature maps. Our idea is that change misses and false detections are due to numerous difficult-to-distinguish samples in the background, while the foreground lacks sufficient discriminative information for interest targets. Therefore, the FS-relation module is designed to improve the associative capacity of interest features, thereby enhancing the discrimination ability for hard case targets. Additionally, to better learn about hard case targets in the background and interest targets in the foreground, we propose an EO-loss for balanced optimization and hard case information learning of foreground-background samples. Finally, we design a multilevel feature fusion module and a lightweight decoder for fusing dual-temporal change information and restoring it to the original pixel space to obtain the final change map.

\subsection{Foreground-Scene Association module}
\subsubsection{Variant feature pyramid network encoder}
To obtain deep features of scene embedding and utilize scene embedding at multiple scales to enhance feature representation, we designed the variant feature pyramid network (V-FPN) for multiscale feature extraction and adaptation to multilayer scene information embedding. The V-FPN consists of a feature extraction branch and a scene information embedding branch. Both branches utilize a variant FPN network divided into multiscale feature layers. The FPN, originally proposed by Liu et al. [6] for object detection tasks, aims to capture strong semantic information in multiple feature layers. To further enhance the model's multiscale feature extraction and fusion capabilities, we designed each layer of the FPN with ResNet as the backbone network for basic feature information extraction. As shown in Figure 2, given a pair of input images $\mathrm{T}^m, m=1,2$, the size of the feature layers in the feature extraction backbone network decreases vertically. This results in a multilevel feature layer set $\mathrm{F}_i^m, i=1,2,3,4 ; m=1,2$, with output strides relative to the input image of (4,8,16,32) pixels. Building on the standard FPN structure, we use feature layers from deep to shallow and lateral feature connections to generate a pyramid feature mapping set $\mathrm{P}_i^m, i=1,2,3,4 ; m=1,2$ with the same dimensions. This method fully links spatial detail information of shallow features with strong semantic information of deep features, aiding in the modelling of multiscale target context information. This process can be expressed by Equation 1.
\begin{equation}
\begin{gathered}
\mathrm{P}_i^m=\tau\left(F_i^m\right)+\sigma\left(\mathrm{P}_{i+1}^m\right), i=1,2,3,4 ; m=1,2
\end{gathered}
\centering
\end{equation}
where $\tau$ represents a learnable 1x1 convolutional layer used for lateral feature connections and $\sigma$ represents a nearest-neighbour upsampling operation with a dimension of 2.

In the final layer $\mathrm{F}_4$ of the pyramid feature set, we designed an important feature flow branch, namely, the foreground-scene information embedding branch. This structure is based on global context aggregation to obtain the spatial scene embedding vector SV, which models the dependency relationship between geographical scenes and interest targets, including interest targets in the foreground and hard case samples. The specific structure and principle of this process will be introduced in the next section. Additionally, in our experiments, we used two types of backbone networks for feature extraction: ResNet\cite{43} and the pyramid vision transformer\cite{44}. The former, with its ingenious residual modules, has strong feature extraction capabilities. Considering accuracy and efficiency, we chose ResNet50 as the first backbone network. Moreover, the PVT can consider both low-level local features and high-level global features. Moreover, the transformer can learn long-range dependencies between features, giving it strong feature representation capabilities at different scales. In the experiments, we used PVTv2 as the second backbone network.

\subsubsection{Foreground-scene associated information embedding}
HRRS is characterized by rich spectral features and complex scene information. Issues such as varying lighting conditions, seasonal changes, and shadows and occlusions caused by tall buildings commonly occur. This means that there is a significant difference in image spectral features and a large intraclass variance in the background, specifically manifested as a large number of hard case samples in the background information, leading to problems such as false positives, false negatives, and pseudochanges in CD. To address this, we propose a foreground-science association module that leverages the scene context of remote sensing space to model the feature representation of foreground targets positively and enhance the feature discrimination ability of background samples. The main principle is shown in Figure 3. The FS-relation module first explicitly models the foreground targets and the geographic scene information and then uses implicit geographic scenes to associate the foreground with the relevant context. This relationship is subsequently used to enhance the input feature mapping, increasing the difference between foreground targets and background hard cases and reducing the likelihood of false alarms and misreporting.
\begin{figure}[ht]
    \centering
    \includegraphics[width=0.5\textwidth]{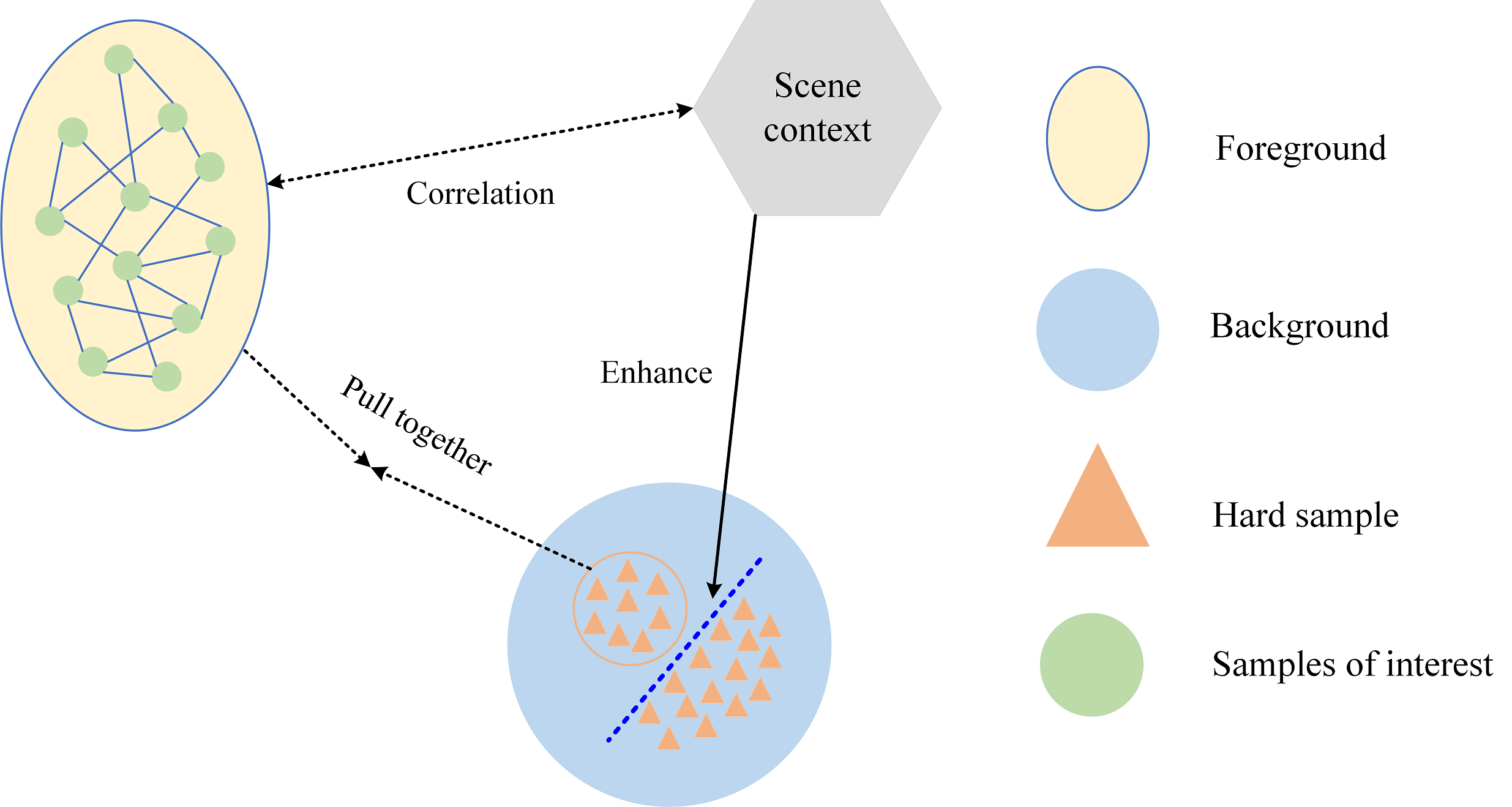}
    \caption{Schematic diagram of the FS-relation module.}
    \label{fig3}
\end{figure}

As depicted in Figure 4, the FS-relation module generates a new feature mapping set $\mathrm{R}_i^m$ based on the feature pyramid collection $\mathrm{P}_i^m, i=1,2,3,4 ; m=1,2$ This module initially re-encodes each layer of $\mathrm{P}_i^m$ to preliminarily form the feature mapping $\mathrm{Q}_i^m$. This mapping is then reweighted according to the correlation map $\mathrm{r}_i^m$, ultimately producing the relation-enhanced feature mapping set $\mathrm{R}_i^m$. The correlation map $\mathrm{r}_i^m$ refers to the similarity matrix between the geographical scene and the foreground representation. To achieve greater discerning $\mathrm{R}_i^m$, the model learns a feature mapping function to align the feature input $\mathrm{P}_i^m$ with the scene vector (SV) into the same dimension, facilitating feature interaction. Here, $Q_i^m \in R^{d \times H \times W}$ is transformed from the feature pyramid layer $P_i^m \in R^{d \times H \times W}$ through the scale projection function $v(\cdot)$, as expressed in Equations 2 and 3.
\begin{equation}
\begin{gathered}
v(\bullet): R^{C \times H \times W} \rightarrow R^{d \times H \times W}
\end{gathered}
\centering
\end{equation}
\begin{equation}
\begin{gathered}
Q_i^m=v\left(P_i^m\right)
\end{gathered}
\centering
\end{equation}
We implement $v(\bullet)$ efficiently, first through a 1×1 convolution layer, followed by batch normalization and ReLU.

The similarity relation matrix set $\mathrm{r}_i^m$ results from the interaction between the scene embedding information and the foreground pyramid features. Therefore, to compute this set, the 1-D scene embedding information $\mathrm{SV} \in \mathrm{R}^d$ interacts with the foreground feature projection $Q_i^m$. Here, the scene embedding information SV is obtained by linearly projecting $P_4^m$, as represented by Equation 4.
\begin{equation}
\begin{gathered}
S V=\varphi\left(P_4^m\right)
\end{gathered}
\centering
\end{equation}
$\varphi(\bullet)$ signifies a scene-focused projection function. For ease of computation, this process is executed by a learnable 1×1 convolution, with the output channel number set to $d$. For the same pair of input images, the remote sensing scene information should be identical, allowing each layer of features in $Q_i^m$ to share SV. At this stage, the similarity relation matrix set $r_i^m$ is derived using Equation 5.
\begin{equation}
\begin{gathered}
r_i^m=\delta\left(S V, Q_i^m\right)=S V \odot Q_i^m
\end{gathered}
\centering
\end{equation}
where $\delta(\bullet)$ denotes the similarity measure function. To simplify the computation and improve the efficiency, the similarity measure is realized by the vector inner product.

The details of the foreground-scene correlation process are illustrated in Figure 4. The final relation-enhanced feature mapping $R_i^m$ is computed according to Equation 6.
\begin{equation}
\begin{gathered}
R_i^m=\frac{\varepsilon\left(P_i^m\right)}{1+\exp \left(-r_i^m\right)}
\end{gathered}
\centering
\end{equation}
$\varepsilon(\bullet)$ is a feature re-encoding structure designed to re-encode the pyramid feature mapping set $P_i^m$, similar to the aforementioned $v(\bullet)$ structure. In this equation, we use a polynomial containing $r_i^m$ to weight and highlight the re-encoded feature mapping $\varepsilon(P_i^m)$.

\begin{figure*}[ht]
    \centering
    \includegraphics[width=6in]{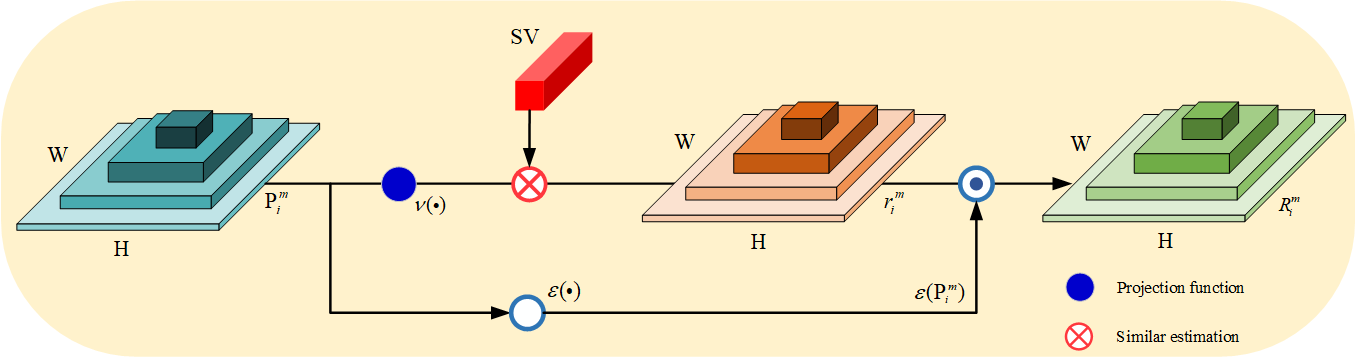}
    \caption{Scene information embedding schematic.}
    \label{fig4}
\end{figure*}

\subsection{Lightweight decoder and dual-branch information fusion}
\begin{figure}[H]
    \centering
    \includegraphics[width=0.5\textwidth]{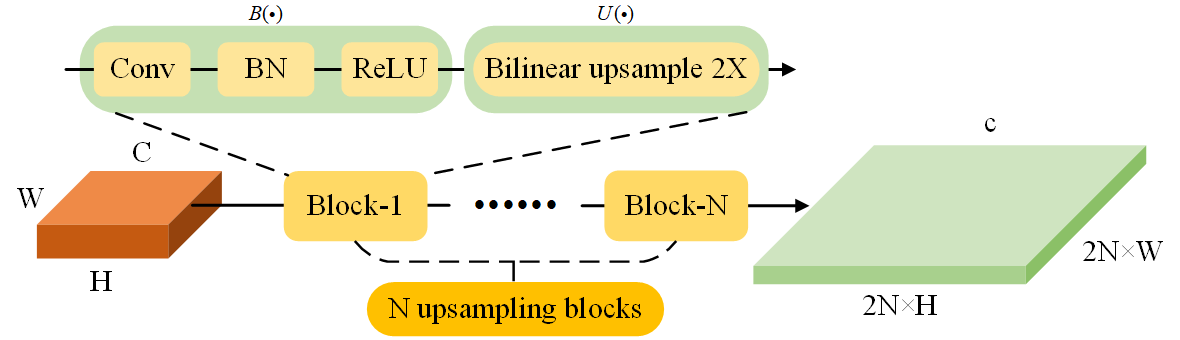}
    \caption{Lightweight decoder architecture diagram}
    \label{fig5}
\end{figure}

As shown in Figure 1, to capture multiscale change information, we fuse the relationship-enhanced feature mapping sets $P_i^1$ and $P_i^2$ of the dual branches, which means that the multiscale feature map $C_i$ can be derived via Equation 7.
\begin{equation}
\begin{gathered}
C_i=a b s\left(R_i^1-R_i^2\right)
\end{gathered}
\centering
\end{equation}
To restore the spatial resolution of the relationship-enhanced feature mapping set, we designed a lightweight decoder, the structure of which is shown in Figure 5. According to the foreground-scene correlation module, given the relationship-enhanced feature mapping $R_i^m$, the upscaled feature mapping set $S_i^m \in R^{c \times \alpha H \times \alpha W}$ can be computed with this decoder. This decoder consists of several upsampling modules, each containing two parts, a dimension transformation operation $B(\bullet)$ and a bilinear interpolation upsampling operation $U(\bullet)$. These are responsible for transforming multiscale feature channels and restoring feature dimensions, respectively, as represented by Equation 8.
\begin{equation}
\begin{gathered}
\hat{S}_i^m=U_N\left(B_N\left(S_i^m\right)\right)
\end{gathered}
\centering
\end{equation}
Here, $N$ represents the number of times the upsampling module is used, which is determined by the dimensions of the $R_i^m$ feature layer. The $B(\bullet)$ operation is achieved through a 3×3 convolution, BN, or ReLU operation, while $U(\bullet)$ represents bilinear interpolation upsampling with a scaling factor of 2; thus, the upsampling scale is $2N$. Subsequently, each layer of the upscaled feature mapping set $\hat{S}_i^m$ is aggregated from high to low to obtain a semantically rich feature mapping $M^m,m=1,2$. This is achieved through pointwise mean computation for feature fusion, followed by a 1×1 convolution for feature integration and parameter computation, as depicted by Equation 9.
\begin{equation}
\begin{gathered}
M^m=\operatorname{Conv}_{1 \times 1}\left(\frac{\sum_{i=1}^4 \hat{S}_i^m}{\max (i)}\right), m=1,2
\end{gathered}
\centering
\end{equation}
For ease of expression and clarity, we collectively refer to the upsampling and aggregation modules as $deo(\bullet)$. Then, the change feature $t$ can be further fused from the dual-branch aggregated features $M^m$, as illustrated by Equation 10.
\begin{equation}
\begin{gathered}
t=a b s\left(M^1-M^2\right)
\end{gathered}
\centering
\end{equation}
Finally, to better represent the multiscale information in the final predictive map, a skip connection is established. The specific operation involves aggregating the change feature $t$ with the multiscale feature information $C_i$, as represented by Equation 11.
\begin{equation}
\begin{gathered}
T=\operatorname{Concat}\left(t, \operatorname{deo}\left(C_i\right)\right)
\end{gathered}
\centering
\end{equation}
We successfully obtained the semantically enriched variational feature $T$. To derive the final predictive change map $T_{\text {change}}$, it is imperative to remap the advanced semantic-laden features $T$   back into the original pixel space. To address this issue, we crafted a lightweight deconvolution decoder based on the fully convolutional network (FCN) framework. This architecture is segmented into three core components: the feature fusion layer, the deconvolution layer, and the classification layer. Each segment is intricately designed, comprising a 3×3 convolution layer with batch normalization (BN) and ReLU, a 4×4 deconvolution layer also equipped with BN and ReLU, and, finally, a deconvolution layer solely with BN. The culmination of this process is the pixel-level classification to discern change information, executed through the sigmoid function, as delineated in Equation 12.
\begin{equation}
\begin{gathered}
T_{\text {change }}=\operatorname{Sigmoid}\left(\eta\left(\operatorname{Conv}_{3 \times 3}(T)\right)\right)
\end{gathered}
\centering
\end{equation}
Here, $\eta(\bullet)$ symbolizes a sequence of two consecutive deconvolution operations.

\subsection{Equilibrium optimization loss function}
Remote sensing imagery backgrounds usually contain a certain number of hard case samples, and models often show poor learning effectiveness for these samples. One reason is that the number of hard case samples is relatively small, and only a portion of the hard case samples related to interest areas are meaningful for model optimization in the later stages of training. Another reason is that in the early stages of training, the model's judgement of hard case samples is uncertain, and directly learning from these samples is less effective. Therefore, we designed an EO-loss to address these issues. The training loss measures the feature distance between the target supervision and the predicted values. The difference between the model's predicted values and the true values for hard case samples is often greater than that for simple samples. Thus, the distribution of hard case samples can be roughly estimated by the loss values. The loss value is positively correlated with the difficulty of sample learning and is generally expressed as follows: the more difficult the sample is to learn, the greater the loss value; the easier the sample is to learn, the smaller the loss value. In the later training stages, the model's predictions for simple samples have already reached a good level, and at this time, only the more challenging samples in the background and foreground are more meaningful for training optimization. Based on this, we propose an EO-loss. As shown in Figure 6, this loss function consists of three steps: hard case sample assessment, dynamic weight optimization, and backpropagation.

To obtain weights that represent the difficulty level of sample data, thereby adjusting the loss distribution pixel by pixel to achieve balanced optimization, we are inspired by the focal loss\cite{35} approach for optimizing hard samples. We use an exponential polynomial with trainable parameters to predict the distribution of hard case samples, which represents the predicted probability and the focus factor. For the pixel-level prediction task of distinguishing between foreground targets and background hard case samples, we aim to adjust the distribution of loss without changing the total value of the loss to avoid the vanishing gradient. Therefore, we introduce a normalization parameter $Z$ to eliminate the impact of outliers and anomalies on the overall results, ensuring stable and efficient learning of hard case samples. This parameter $Z$ must ensure that Equation 13 is valid.
\begin{equation}
\begin{gathered}
\sum_{i=0}^{H \times W} \operatorname{loss}\left(p_i, y_i\right)=\frac{1}{Z} \sum_{i=0}^{H \times W}\left(1-p_i\right)^\gamma \operatorname{loss}\left(p_i, y_i\right)
\end{gathered}
\centering
\end{equation}
Here, ${loss}\left(p_i, y_i\right)$ represents the binary cross-entropy loss value for the i-th pixel, which can be calculated from the predicted probability value $p_i$ and the actual value $y_i$. Therefore, the weight of each pixel's loss value is $\frac{1}{Z}(1-p)^\gamma$.
\begin{figure}[ht]
    \centering
    \includegraphics[width=0.5\textwidth]{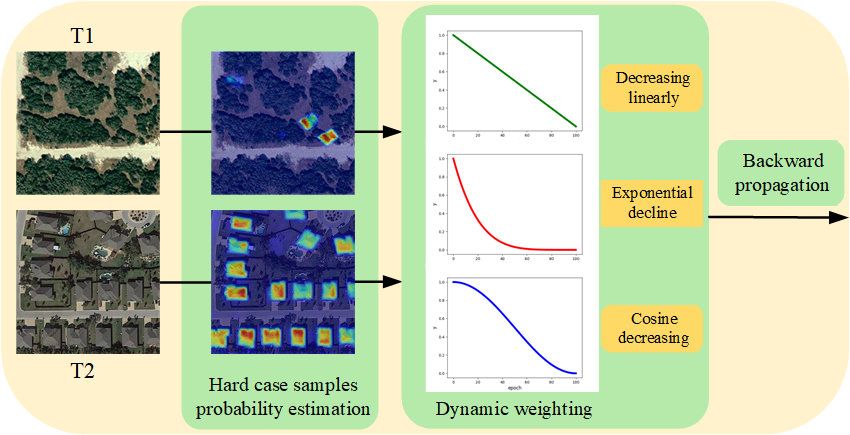}
    \caption{Equilibrium optimization loss function structure.}
    \label{fig6}
\end{figure}
The learning effectiveness of hard case samples depends on the optimization method of the model. In the early stages of training, the model's confidence in sample category determination is not high, and dealing with too many hard case samples at this time may lead to an unstable learning process. To ensure stable and convergent model training, we propose a dynamic weighting strategy based on a decay function. This strategy focuses on learning from simple samples in the early training stages and strengthens the optimization of hard case samples in the later stages. Based on the binary cross-entropy loss $BCE(\bullet)$, we can express the EO-loss as shown in Equation 14.
\begin{equation}
\begin{gathered}
loss=\left[\lambda(t)\left(1-\frac{1}{Z}\left(1-p_i\right)^\gamma\right)+\frac{1}{Z}\left(1-p_i\right)^\gamma\right] BCE\left(p_i, y_i\right)
\end{gathered}
\centering
\end{equation}
Here, $\lambda(t)$ represents a weight function related to the training process $t$, which is used for weighting nonhard case samples, and $\lambda(t) \in(0,1)$ decreases monotonically. Therefore, as training progresses, the model's confidence in predicting nonhard case samples increases continuously, and the focus of the loss value distribution gradually shifts to hard case samples. Considering that there are multiple methods to decrease $\lambda(t)$, we design three representative and computationally convenient weight functions for weighting nonhard case samples, with specific details shown in TABLE I. A linear decrease is used for the model to transition smoothly from simple to hard case samples; an exponential decrease is based on the idea that the learning process of simple samples progresses very quickly, hence further shifting the optimization focus towards hard case samples and increasing the time for optimizing hard case samples; and a cosine decrease distributes the optimization weights evenly between simple and hard case samples, reducing the proportion of the transition process.
\begin{table}[]
\centering
\caption{INFORMATION ON DIFFERENT WEIGHTING FUNCTIONS}
\label{TABLE I}
\begin{tabular}{@{}ccc@{}}
\toprule
Function    & Expression & hyperparameter \\ \midrule
Linear      & $\lambda(t)=1-\frac{t}{\text { step }}$           & step           \\

Exponential & $\lambda(t)=\left(1-\frac{t}{\text { step }}\right)^{\text {decaty }}$           & step and decay \\

Cosine      & $\lambda(t)=\frac{1}{2}\left(1+\cos \left(\frac{t \pi}{s t e p}\right)\right)$           & step           \\ \bottomrule
\end{tabular}
\end{table}

\section{Experiments and discussion}
To evaluate the performance of HSONet, we conducted experiments on four public CD datasets, LEVIR-CD, Google-CD, CDD, and SYSU-CD, and compared them with the latest RS-CD methods. Additionally, we designed ablation studies and effectiveness experiments and discussed the experimental results.
\subsection{Datasets}
\subsubsection{LEVIR-CD[17]}
A large-scale building change detection dataset consisting of 637 pairs of Google Earth images, each with a size of 1024×1024 pixels and a resolution of 0.5 m, spanning from 2002 to 2018. It contains a variety of building change information, such as villas, large factories, high-rise apartments, and garages.
\subsubsection{Google-CD[18]}
This dataset records changes in various types of buildings, including large factories, villages, and warehouses, in the suburbs of Guangzhou, China, from 2006 to 2019. The 19 pairs of images are sourced from Google Earth, with spatial resolutions of 0.55 m and image sizes ranging from 1006×1168 to 4936×5224.
\subsubsection{CDD[16]}
This dataset contains 11 pairs of multisource remote sensing images collected in different seasons; 7 pairs have a resolution of 7425×2202 pixels, and 4 pairs have 1900×1000 pixels, with resolutions varying from 0.03 m to 1 m. The challenge of this dataset lies in accurately detecting changes in buildings, roads, farmland, and vehicles, regardless of the impact of seasonal changes.
\subsubsection{SYSU-CD[19]}
This dataset includes 20,000 pairs of aerial images collected in Hong Kong, China, from 2007 to 2014; all the images have a size of 256×256 pixels and a resolution of 0.5 m. The main types of changes in the data include urban buildings, road construction, marine construction, and vegetation changes.

\subsection{Metrics}
\subsubsection{Evaluation criteria}
To evaluate the performance of the model, we used seven evaluation metrics for result presentation: precision (P), recall (R), F1 score (F1), intersection over union (IOU), mean intersection over union (mIOU), overall accuracy (OA), and kappa coefficient (Kappa). In the RS-CD tasks, higher precision indicates more correct detections in positive cases, higher recall means fewer losses in predicted positive cases, and larger F1, IOU, and mIOU indicate better CD performance. A higher kappa coefficient indicates better consistency between two sets of predictions, while the OA is an overall assessment of all the predictions being correctly classified. The calculation formulas for these metrics are as follows.
\begin{equation}
\begin{gathered}
P=\frac{TP}{TP+FP}
\end{gathered}
\centering
\end{equation}
\begin{equation}
\begin{gathered}
R=\frac{TP}{TP+FN}
\end{gathered}
\centering
\end{equation}
\begin{equation}
\begin{gathered}
F 1=\frac{1}{\mathrm{P}^{-1}+\mathrm{R}^{-1}}
\end{gathered}
\centering
\end{equation}
\begin{equation}
\begin{gathered}
IOU=\frac{TP}{TP+FN+FP}
\end{gathered}
\centering
\end{equation}
\begin{equation}
\begin{gathered}
OA=\frac{TP+TN}{TP+FN+FP+TN}
\end{gathered}
\centering
\end{equation}
\begin{equation}
\begin{gathered}
m I O U=\frac{1}{2}\left(\frac{T P}{T P+F N+F P}+\frac{T N}{T N+F N+F P}\right)
\end{gathered}
\centering
\end{equation}
\begin{equation}
\begin{gathered}
K a p p a=\frac{O A-P e}{1-P}
\end{gathered}
\centering
\end{equation}

\subsubsection{Implementation details}
We conducted experiments using an NVIDIA 4080Ti graphics card and trained the model with a minibatch Adam optimizer. The PyTorch deep learning framework was used. The Epoch was set to 150, batch size to 16, learning rate to 5×10-4, weight decay to5×10-4, Step size to 50, and momentum to 0.9. For the CDD dataset, we cropped the original images into small images of 256×256 pixels and randomly performed simple operations such as rotation, flipping, and central cropping. We obtained 10,000/3,000/3,000 pairs of image patches for training, validation, and testing, respectively. For the LEVIR-CD dataset, we cropped the original images into nonoverlapping blocks of 256×256 pixels, similarly performing random rotations, flipping, and central cropping for data augmentation. We divided the images randomly into three parts: 8392/600/1200 pairs for training, validation, and testing. For the Google-CD dataset, we cropped the original images into blocks of 256×256 pixels and performed random rotations, flipping, and central cropping steps; we obtained 2661/157/312 pairs of images for training, validation, and testing. For the SYSU-CD dataset, we cropped the original images into small images of 256×256 pixels, similarly performing random rotations, flipping, and central cropping, and obtained 14,001/1,999/4,000 pairs of images for training, validation, and testing.

\subsection{Results and Discussion}
\subsubsection{Benchmark methods}
In this subsection, we introduce the benchmark methods used for comparison. To demonstrate the superiority of HSONet, we selected the most classic research achievements in each subdirection for comparison, including three pure convolution-based CD methods, FC-EF, FC-Siam-Di, and FC-Siam-Co; a method based on channel and spatial attention, DASNet; a method embedding ViT into the CNN backbone BiT; a pure Swin-T based method, SwinSUNet; a method based on cross-temporal joint attention, DMINet; a method based on cross-interaction attention and multiscale feature fusion, ICIF-Net; a CD method for building edge feature enhancement, EGCTNet; a method based on high-frequency information enhancement and spatial attention, HFA-Net; and a 3-D attention method, USSFC-Net, based on spatial-spectral feature synergy.

\begin{enumerate}
\item[(a)] FC-EF\cite{22}: An early feature fusion CD method. It directly concatenates dual-temporal images in the channel dimension before feeding them into the network.

\item[(b)] FC-Siam-Co\cite{22}: A weight-sharing dual-encoder Siamese network that acquires change information by fusing features at various levels.

\item[(c)] FC-Siam-Di\cite{22}: Compared with FC-Siam-Co, this method changes the feature map fusion method in the network from concatenation to absolute difference.

\item[(d)] DASNet\cite{11}: Based on VGG16 or ResNet50, this Siamese network introduces spatial and channel attention in the feature encoding phase to enhance the network's focus on interest changes and resist pseudochange interference. The change map is obtained through a feature metric module.

\item[(e)] BIT\cite{29}: This method embeds a transformer module in the convolutional network, converting dual-temporal feature mappings into compact semantic tokens to facilitate modelling of the spatiotemporal context in feature space, thereby optimizing feature representation in image space.

\item[(f)] SwinSUNet\cite{34}: Based on Swin-T, this dual U-shaped CD network uses multiple pure transformer encoders to capture the spatiotemporal context of dual-temporal images. After feature fusion and linear mapping, the change information is decoded using multiple Swin-T blocks combined with skip connections.

\item[(g)] DMINet\cite{13}: Based on dual-branch ResNet18 for feature extraction, DMINet unifies self-attention and cross-attention in one module to guide the global feature distribution of each input. Two change information acquisition structures are also designed, namely, the fusion method based on subtraction and concatenation and a multilevel differential aggregation method based on incremental feature alignment.

\item[(h)] ICIFNet\cite{30}: This method designs a four-branch cross-interaction feature extraction structure with parallel CNN and ViT to promote the mutual penetration of local and global features. Mask-based aggregation and spatial alignment (SA) schemes are also introduced for scale integration, achieving information integration at different resolutions.

\item[(i)] EGCTNet\cite{48}: Focusing on the issue that local fine features and global information in CD cannot be obtained simultaneously, a fusion encoder is designed combining a CNN and transformers. Additionally, an edge detection branch is proposed that uses edge information to guide mask feature generation.

\item[(j)] HFA-Net\cite{49}: Addressing the issue of insufficient high-frequency information acquisition for clearly defined targets, a high-frequency attention-guided module is proposed. It consists of two main stages: first, spatial attention is used to search for and focus on buildings; second, high-frequency enhancement is used to highlight the high-frequency information of features, which better represents the edges of changed buildings.

\item[(k)] USSFC-Net\cite{20}: Simultaneously modelling spatial and spectral features, this method proposes an efficient collaborative network to generate three-dimensional attention for richer feature information. Additionally, a multiscale decoupled convolution is designed to flexibly capture the multiscale features of changing objects.
\end{enumerate}

\subsubsection{Performance comparison}
\begin{figure}[ht]
    \centering
    \includegraphics[width=0.5\textwidth]{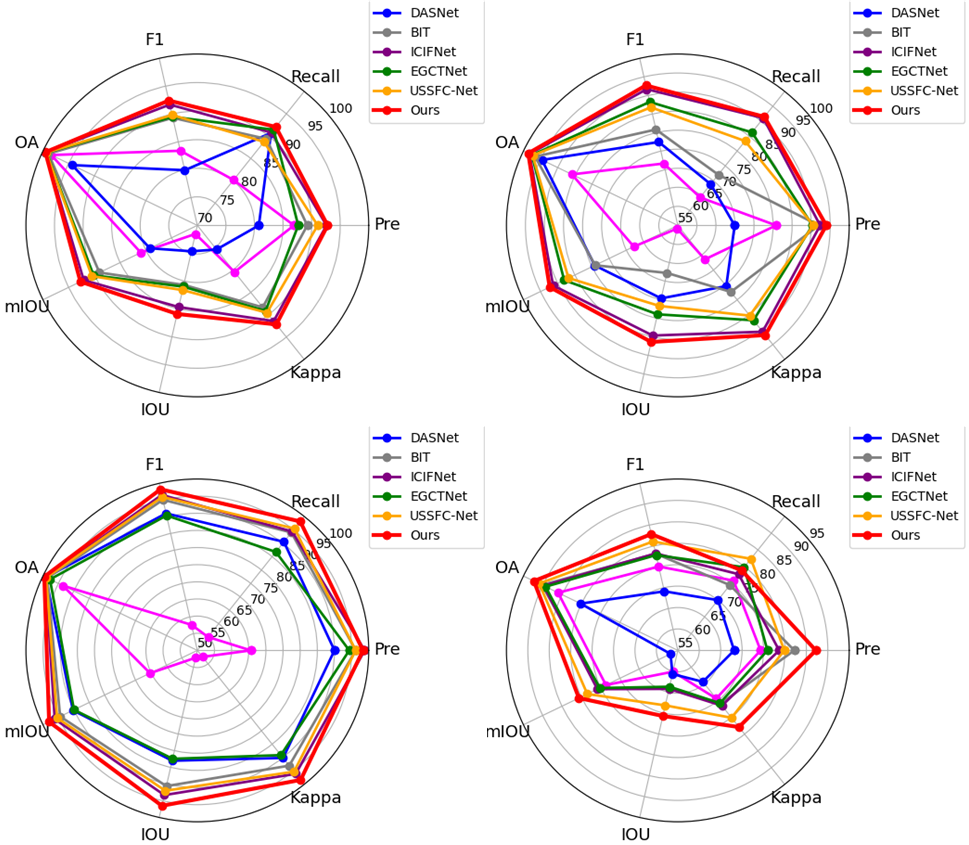}
    \caption{Precision metric radar chart. The above figures show the accuracy metric radar charts on LEVIR-CD, Google-CD, CDD, and SYSU-CD.}
    \label{fig7}
\end{figure}
To accurately assess the effectiveness of HSONet, we conducted quantitative experiments on four public CD datasets. TABLE II, TABLE III, TABLE IV, and TABLE V show the accuracy of HSONet on the test sets of LEVIR-CD, Google-CD, CDD, and SYSU-CD, respectively. The quantitative results indicate that our method consistently outperforms the others on three datasets and demonstrates significant advantages on the SYSU-CD dataset. For instance, on the first three datasets, HSONet outperforms the other comparative methods in most accuracy metrics, with F1 scores exceeding those of the latest method, DMINet, by 1.16\%, 1.15\%, and 1.72\%, respectively. Objectively speaking, our method has achieved certain breakthroughs in all seven accuracy metrics on all four datasets, proving the significance of introducing an equilibrium optimization loss function and a novel CD network architecture with a foreground-science association module. Additionally, Figures 8, 9, 10, and 11 show the visualization results of HSONet on each test set. To clearly observe the accuracy of each area classification, we used different colours to represent TP (blue), TN (light blue), FP (red), and FN (orange). To obtain a clearer view of the performance comparison between the various methods, we visualized the scores of the seven accuracy metrics using radar charts; as Figure 7 shows, our method has several advantages in all the metrics on the four datasets, and the homogeneity is good, with no poor accuracy results.
\paragraph{Experimental Results on the LEVIR-CD Dataset}
As Figure 8 shows, we selected typical samples from the LEVIR-CD test set for visual comparison, such as the shadow area in the middle of the ring-shaped building in (1), the large scene changes in (2), and the strip-like small target buildings in (3). For (1), only our method shows strong robustness to shadow areas and more reasonable segmentation results for the interior and boundaries of buildings. In (2), HSONet exhibits stronger resistance to noninteresting changes and fewer misses and false detections than do the other methods. In (3), our method not only accurately detects the complete change area of large buildings but also senses changes in small targets at the bottom right, which were not annotated in the GT, demonstrating HSONet's strong perception of detailed information. In summary, our method achieves SOTA performance according to the qualitative results on LEVIR-CD, which is consistent with the quantitative results in TABLE II.
\begin{table}[]
\centering
\caption{COMPARISON RESULT ON LEVIR-CD TEST SET}
\label{TABLE }
\resizebox{\columnwidth}{!}{%
\begin{tabular}{@{}ccccccccc@{}}
\toprule
Method     & Year & Precision      & Recall         & F1             & OA             & mIOU           & IOU            & Kappa          \\ \midrule
FC-EF      & 2018 & 86.91          & 80.17          & 83.40          & 98.39          & 80.98          & 71.53          & 80.44          \\
FC-Siam-Di & 2018 & 89.53          & 83.31          & 86.31          & 98.67          & 83.87          & 75.92          & 84.28          \\
FC-Siam-Co & 2018 & 91.99          & 76.77          & 83.69          & 98.49          & 81.42          & 71.96          & 80.94          \\
DASNet     & 2020 & 80.76          & 90.70          & 79.91          & 94.32          & 79.22          & 74.65          & 75.39          \\
BIT        & 2021 & 89.24          & 89.37          & 89.31          & 98.92          & 89.02          & 80.68          & 88.35          \\
SwinSUNet  & 2022 & 88.03          & 84.76          & 86.37          & 99.16          & 87.57          & 76.01          & 85.93          \\
DMINet     & 2022 & \textbf{93.02} & 89.58          & 91.26          & 99.46          & 91.69          & 83.94          & 90.99          \\
ICIFNet    & 2022 & 92.61          & 90.80          & 91.69          & 99.48          & 92.07          & 84.66          & 91.43          \\
EGCTNet    & 2022 & 87.66          & 91.42          & 89.50          & 99.33          & 90.15          & 81.00          & 89.15          \\
HFA-Net    & 2022 & 90.10          & 80.48          & 85.02          & 99.11          & 86.51          & 73.94          & 94.56          \\
USSFC-Net  & 2023 & 91.06          & 88.72          & 89.81          & 99.37          & 90.48          & 81.61          & 89.55          \\
Ours       & -    & 92.80          & \textbf{92.04} & \textbf{92.42} & \textbf{99.53} & \textbf{92.71} & \textbf{85.90} & \textbf{92.17} \\ \bottomrule
\end{tabular}%
}
\end{table}
\begin{table}[]
\centering
\caption{COMPARISON RESULT ON Google-CD TEST SET}
\label{TABLE }
\resizebox{\columnwidth}{!}{%
\begin{tabular}{@{}ccccccccc@{}}
\toprule
Method     & Year & Precision      & Recall         & F1             & OA             & mIOU           & IOU            & Kappa          \\ \midrule
FC-EF      & 2018 & 80.81          & 64.39          & 71.67          & 85.85          & 67.83          & 55.85          & 66.40          \\
FC-Siam-Di & 2018 & 85.44          & 63.28          & 72.71          & 87.27          & 69.37          & 57.12          & 70.01          \\
FC-Siam-Co & 2018 & 82.07          & 64.73          & 72.38          & 84.56          & 68.39          & 56.71          & 69.05          \\
DASNet     & 2020 & 69.92          & 68.83          & 77.46          & 94.59          & 79.22          & 74.65          & 75.39          \\
BIT        & 2021 & 92.04          & 72.03          & 80.82          & 96.59          & 79.11          & 67.81          & 77.33          \\
SwinSUNet  & 2022 & 90.77          & 75.92          & 82.68          & 96.59          & 83.39          & 70.48          & 80.81          \\
DMINet     & 2022 & 93.06          & 90.92          & 91.98          & 98.30          & 91.63          & 85.15          & 91.03          \\
ICIFNet    & 2022 & 92.37          & 91.02          & 91.69          & 98.23          & 91.35          & 84.66          & 90.70          \\
EGCTNet    & 2022 & 90.34          & 86.23          & 88.24          & 97.53          & 88.12          & 78.95          & 86.86          \\
HFA-Net    & 2022 & 81.97          & 77.91          & 79.89          & 95.80          & 80.97          & 66.52          & 77.55          \\
USSFC-Net  & 2023 & 90.58          & 83.33          & 86.80          & 97.29          & 86.86          & 76.69          & 85.3           \\
Ours       & -    & \textbf{95.13} & \textbf{91.25} & \textbf{93.14} & \textbf{98.56} & \textbf{92.79} & \textbf{87.17} & \textbf{92.34} \\ \bottomrule
\end{tabular}%
}
\end{table}
\begin{table}[]
\centering
\caption{COMPARISON RESULT ON CDD TEST SET}
\label{TABLE }
\resizebox{\columnwidth}{!}{%
\begin{tabular}{@{}ccccccccc@{}}
\toprule
Method     & Year & Precision      & Recall         & F1             & OA             & mIOU           & IOU            & Kappa          \\ \midrule
FC-EF      & 2018 & 65.60          & 55.01          & 57.65          & 93.58          & 65.32          & 52.20          & 52.42          \\
FC-Siam-Di & 2018 & 78.25          & 65.76          & 69.04          & 94.44          & 67.24          & 56.57          & 62.39          \\
FC-Siam-Co & 2018 & 74.51          & 73.87          & 71.16          & 94.92          & 69.48          & 60.11          & 66.44          \\
DASNet     & 2020 & 90.00          & 90.50          & 91.00          & 99.10          & 90.28          & 83.01          & 90.12          \\
BIT        & 2021 & 96.19          & 93.99          & 95.07          & 98.62          & 94.43          & 90.61          & 93.00          \\
SwinSUNet  & 2022 & 95.70          & 92.30          & 94.00          & 98.50          & 93.36          & 91.74          & 93.28          \\
DMINet     & 2022 & 98.40          & 94.66          & 96.49          & 99.15          & 96.13          & 93.23          & 96.01          \\
ICIFNet    & 2022 & \textbf{98.81} & 94.35          & 96.53          & 99.16          & 96.17          & 93.29          & 96.05          \\
EGCTNet    & 2022 & 94.32          & 86.74          & 90.37          & 97.72          & 89.94          & 82.43          & 89.08          \\
HFA-Net    & 2022 & 92.23          & 62.55          & 74.54          & 94.73          & 76.85          & 59.42          & 71.72          \\
USSFC-Net  & 2023 & 96.19          & 95.46          & 95.92          & 98.97          & 95.41          & 91.99          & 95.24          \\
Ours       & -    & 98.32          & \textbf{98.12} & \textbf{98.21} & \textbf{99.56} & \textbf{98.00} & \textbf{96.49} & \textbf{98.33} \\ \bottomrule
\end{tabular}%
}
\end{table}
\begin{table}[]
\centering
\caption{COMPARISON RESULT ON SYSU-CD TEST SET}
\label{TABLE }
\resizebox{\columnwidth}{!}{%
\begin{tabular}{@{}ccccccccc@{}}
\toprule
Method     & Year & Precision      & Recall         & F1             & OA             & mIOU           & IOU            & Kappa          \\ \midrule
FC-EF      & 2018 & 74.32          & 75.84          & 75.07          & 86.02          & 73.62          & 60.09          & 69.37          \\
FC-Siam-Di & 2018 & 89.13          & 61.21          & 72.57          & 82.11          & 68.77          & 56.96          & 67.01          \\
FC-Siam-Co & 2018 & 82.54          & 71.03          & 76.35          & 86.17          & 74.21          & 61.75          & 70.09          \\
DASNet     & 2020 & 68.14          & 70.01          & 69.14          & 80.16          & 56.88          & 60.65          & 64.37          \\
BIT        & 2021 & 82.18          & 74.49          & 78.15          & 90.18          & 75.29          & 64.13          & 70.93          \\
SwinSUNet  & 2022 & 78.27          & 73.93          & 76.04          & 89.01          & 74.02          & 61.34          & 68.92          \\
DMINet     & 2022 & 82.08          & \textbf{84.86} & \textbf{83.45} & 92.06          & 80.53          & \textbf{71.60} & 78.23          \\
ICIFNet    & 2022 & 78.64          & 77.75          & 78.20          & 89.77          & 75.84          & 64.21          & 71.51          \\
EGCTNet    & 2022 & 75.92          & 79.81          & 77.82          & 89.27          & 75.24          & 63.69          & 70.75          \\
HFA-Net    & 2022 & 80.93          & 70.91          & 75.60          & 89.20          & 73.90          & 60.76          & 68.70          \\
USSFC-Net  & 2023 & 79.84          & 82.34          & 81.07          & 90.93          & 78.45          & 68.16          & 75.11          \\
Ours       & -    & \textbf{87.21} & 78.88          & 82.84          & \textbf{92.29} & \textbf{80.62} & 70.70          & \textbf{78.28} \\ \bottomrule
\end{tabular}%
}
\end{table}
\paragraph{Experimental Results on the Google-CD Dataset}
Similarly, we selected representative prediction samples from Google-CD for visual comparison. As shown in Figure 9, large scene changes occur in (1), (2), and (3). Specifically, (1) involves interference from noninteresting changes such as roads and trees, and the challenge in (2) is how to reduce misclassifications around buildings. Despite the limited number of samples in this dataset, the DMI and our method have significant advantages in terms of detection accuracy and completeness compared to the other methods, consistent with the quantitative results in TABLE III. In addition, only HSONet simultaneously achieves high recognition rates and low false alarm rates for all three datasets, further demonstrating the superior performance of our method on Google-CD images.
\begin{figure*}[ht]
    \centering
    \includegraphics[width=6in]{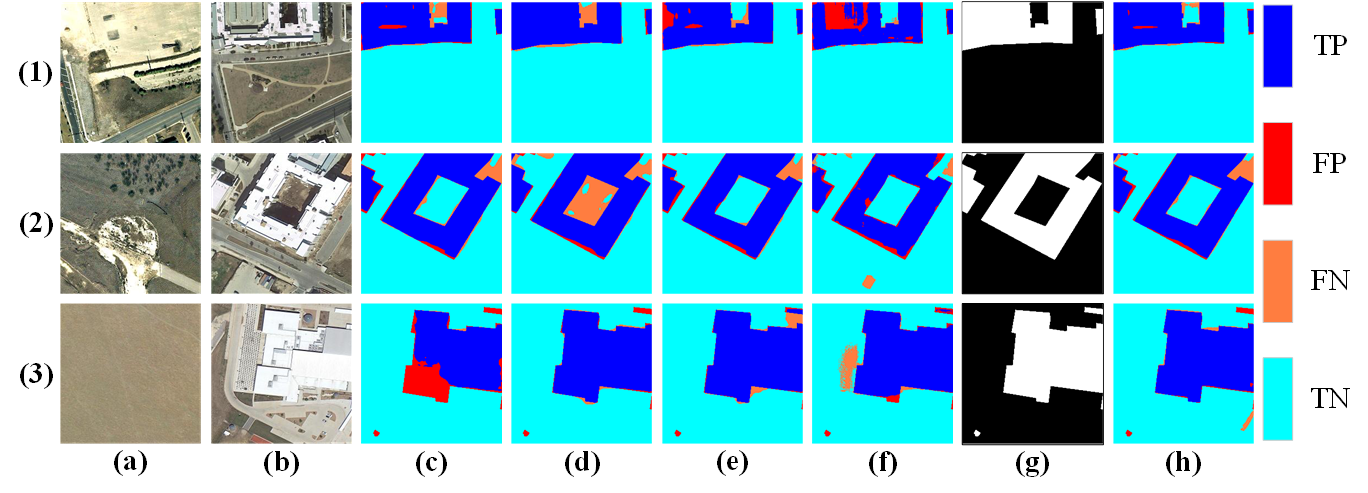}
    \caption{Visualization results of several CD methods on the LEVIR-CD test set. (a)T1, (b)T2, (c)DMINet, (d)EGCTNet, (e)ICIFNet, (f)USSFC-Net, (g)GT, and (h)Ours, where blue, red, orange, and light blue denote TP, FP, FN and TN, respectively.}
    \label{fig8}
\end{figure*}
\begin{figure*}[ht]
    \centering
    \includegraphics[width=6in]{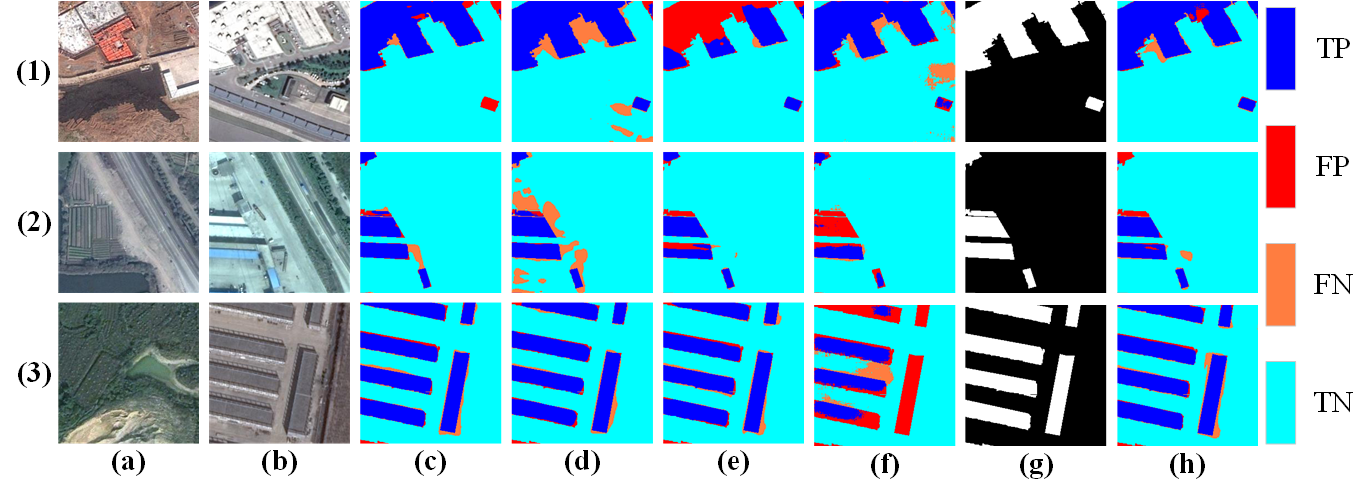}
    \caption{Visualization results of several CD methods on the Google-CD test set. (a)T1, (b)T2, (c)DMINet, (d)EGCTNet, (e)ICIFNet, (f)USSFC-Net, (g)GT, and (h)Ours, where blue, red, orange, and light blue denote TP, FP, FN and TN, respectively.}
    \label{fig9}
\end{figure*}
\begin{figure*}[ht]
    \centering
    \includegraphics[width=6in]{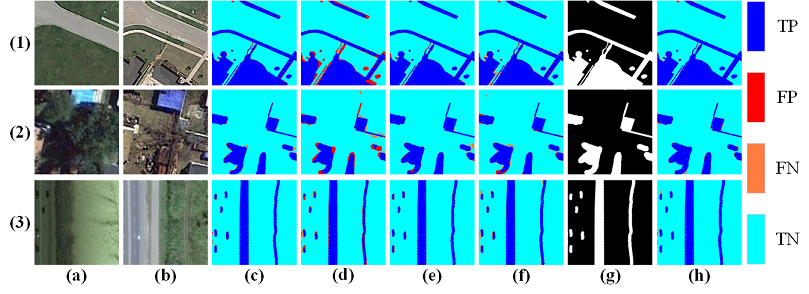}
    \caption{Visualization results of several CD methods on the CDD test set. (a)T1, (b)T2, (c)DMINet, (d)EGCTNet, (e)ICIFNet, (f)USSFC-Net, (g)GT, and (h)Ours, where blue, red, orange, and light blue denote TP, FP, FN and TN, respectively.}
    \label{fig10}
\end{figure*}
\begin{figure*}[ht]
    \centering
    \includegraphics[width=6in]{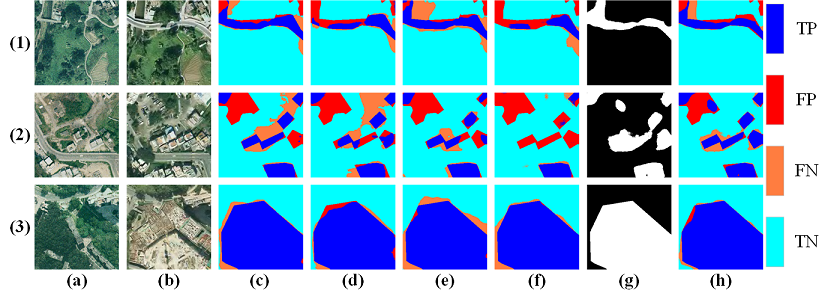}
    \caption{Visualization results of several CD methods on the SYSU-CD test set. (a)T1, (b)T2, (c)DMINet, (d)EGCTNet, (e)ICIFNet, (f)USSFC-Net, (g)GT, and (h)Ours, where blue, red, orange, and light blue denote TP, FP, FN and TN, respectively.}
    \label{fig11}
\end{figure*}
\paragraph{Experimental Results on the CDD Dataset}
As Figure 10 shows, we selected typical samples from the CDD test set for visual comparison. These include (1) numerous small target changes, (2) challenges due to tree occlusion affecting feature information discrimination, and (3) significant seasonal changes, which can interfere with the change detection of rural roads. As observed in TABLE IV, HSONet consistently achieves the highest recognition scores with the least FPs and FNs, surpassing the other comparative methods. This indicates that our method can accurately detect changes in interest under seasonal interference, perceive small target changes, rarely miss or falsely detect changes, and is less affected by light, colour, and weather.
\paragraph{Experimental Results on the SYSU-CD Dataset}
While the first three datasets mainly focus on changes in buildings, to verify our method's detection performance for different types of target changes, we selected typical samples from the SYSU-CD test set for visual comparison. As shown in Figure 11, HSONet yields fewer misjudgments and higher recognition accuracy when detecting changes in ambiguous areas such as roads, bare land, and residential areas. It also shows robustness and resistance to noninteresting changes such as forests and rural roads. This proves that, compared to other methods, our method has a universal ability to discriminate changes in interest and can more accurately detect changes in various categories of land objects, consistent with the quantitative results in TABLE V.

\subsubsection{Learning curve comparison}
To evaluate the performance of HSONet, we compared the F1 score variation over epochs for USSFC-Net, EGCTNet, and HSONet on four datasets. Figure 12 shows the change line graphs for LEVIR-CD, Google-CD, CDD, and SYSU-CD from top to bottom and left to right. The graph shows that our model has higher accuracy, faster convergence, and a more stable training process than the other two methods, with very few large fluctuations in accuracy during training. This indicates that HSONet is more robust and stable than the other models and has stronger robustness. Additionally, the model converges after reaching a certain point, and training for more epochs does not significantly enhance the CD capability. Therefore, in our experiments, we set the epochs according to the dataset: 60 epochs for LEVIR-CD and 100 epochs for Google-CD, CDD, and SYSU-CD.
\begin{figure}[ht]
    \centering
    \includegraphics[width=0.5\textwidth]{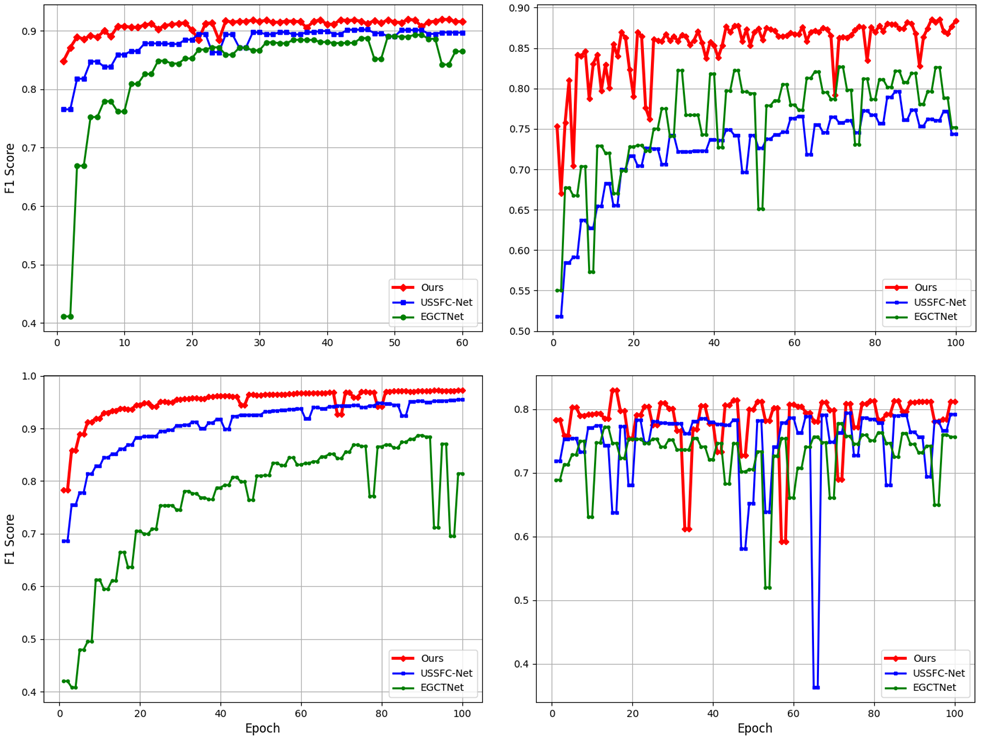}
    \caption{F1 score compare with USSFC-Net and EGCTNet on LEVIR-CD, Google-CD, CDD, and SYSU-CD validation set.}
    \label{fig12}
\end{figure}

\subsection{Ablation Study}
\begin{table*}
\centering
\caption{COMPARISON OF RESULTS WITH DIFFERENT LOSS FUNCTIONS}
\begin{tabular}{cccccccccc} 
\hline
Loss                     &    & \multicolumn{2}{c}{Backbone} & \multicolumn{2}{c}{LEVIR-CD}    & \multicolumn{2}{c}{Google-CD}   & \multicolumn{2}{c}{SYSU-CD}      \\ 
\cmidrule{3-10}
function                 & FS & PVT & ResNet                 & F1             & IOU            & F1             & IOU            & F1             & IOU             \\ 
\midrule
\multirow{3}{*}{BCE}     &    &     & \checkmark                      & 91.74          & 84.74          & 91.40          & 84.18          & 76.65          & 62.14           \\
                         & \checkmark  & \checkmark   &                        & \textbf{92.24} & \textbf{85.59} & \textbf{92.01} & \textbf{85.38} & \textbf{82.72} & \textbf{70.58}  \\
                         & \checkmark  &     & \checkmark                      & 92.12          & 85.39          & 91.90          & 85.01          & 81.42          & 68.66           \\ 
\hline
\multirow{3}{*}{SF-loss} &    &     & \checkmark                      & 91.77          & 84.80          & 90.87          & 83.27          & 73.39          & 57.96           \\
                         & \checkmark  & \checkmark   &                        & \textbf{92.31} & \textbf{85.73} & \textbf{92.16} & \textbf{85.46} & \textbf{82.53} & \textbf{70.25}  \\
                         & \checkmark  &     & \checkmark                      & 91.86          & 84.89          & 91.20          & 83.82          & 82.14          & 69.69           \\ 
\hline
\multirow{3}{*}{EO-loss} &    &     & \checkmark                      & 91.75          & 84.76          & 91.40          & 84.17          & 74.73          & 59.65           \\
                         & \checkmark  & \checkmark   &                        & \textbf{92.42} & \textbf{85.90} & \textbf{93.14} & \textbf{87.17} & \textbf{82.84} & \textbf{70.70}  \\
                         & \checkmark  &     & \checkmark                      & 91.87          & 84.97          & 91.69          & 84.65          & 81.96          & 69.44           \\
\hline
\end{tabular}
\end{table*}
\subsubsection{Comparison of different loss function effects}
To explore the contribution of EO-loss to model performance, we compared the original loss function with the EO-loss. As shown in TABLE VI, in the control experiments using the PVT, the accuracy of the EO-loss on the three datasets is 0.18\%, 1.13\%, and 0.12\% greater than that of the BCE, and 0.11\%, 0.98\%, and 0.31\% greater than that of the SF-loss. These results indicate that EO-loss indeed has certain advantages. However, since hard case samples constitute a smaller proportion of the data samples, the improvement in accuracy is not very large. This finding demonstrates the effectiveness and rationality of the method.

\subsubsection{Comparison of the FS-relation effect}
To explore the contribution of the FS-relation module to the performance of HSONet, we conducted ablation experiments on the FS-relation module on three datasets. As shown in TABLE VI, when ResNet was used as the backbone network, the three sets of experiments introducing the FS-relation module were comprehensively superior to those without it. For example, when the loss function is BCE, the F1 scores of the former are 0.38\%, 0.50\%, and 4.77\% greater than those of the latter on three datasets, and similar results are obtained for the SF-loss and EO-loss experiment sets. This finding suggests that enhancing the feature representation of hard case targets by modelling the association of foreground targets of interest with related context in latent remote sensing spatial scene information is a feasible feature modelling strategy.

\subsubsection{Comparison of different backbone effects}
To verify the role of the multiscale structure-designed backbone in the network, we compared the experimental results of PVT and ResNet on three datasets. As shown in TABLE VI, in all six sets of comparative experiments, PVT achieves more competitive accuracy results than ResNet. For instance, when the loss function is EO-loss, the F1 scores of the PVT group are 0.55\%, 1.45\%, and 0.88\% greater than those of the ResNet group. This indicates that the PVT has a stronger feature extraction capability than ResNet and is more suitable for CD tasks. This might be due to ViT's superior global modelling capability and the multiscale information of the pyramid feature layer, which allows the FS-relation module to obtain more representative scene embedding vector SVs, thereby enhancing the feature representation of background hard case targets.

\subsection{Parameter Verification Experiment}
There are many important parameters and functions in the proposed network, and to explore the effectiveness of these parameters and functions on model performance, we conduct parameter validation experiments on three datasets.
\subsubsection{Effect of the dynamic weighting function}
\begin{table*}
\centering
\caption{COMPARISON OF RESULTS FOR WEIGHTING FUNCTIONS}
\begin{tabular}{ccccccccccc} 
\hline
                                &          &            &                 &            & \multicolumn{2}{c}{LEVIR-CD}    & \multicolumn{2}{c}{Google-CD}   & \multicolumn{2}{c}{CDD}          \\ 
\cline{6-11}
~
  Backbone                    & ~
  Norm & ~
  Linear & ~
  Exponential & ~
  Cosine & F1             & IOU            & F1             & IOU            & F1             & IOU             \\ 
\hline
\multirow{5}{*}{~
  ~
  PVT}    & ~        & ~          & ~               & ~          & 92.08          & 85.32          & 91.90          & 85.01          & 97.20          & 94.61           \\
                                & \checkmark        & ~          & ~               & ~          & 92.31          & 85.72          & 92.15          & 85.45          & 97.38          & 94.88           \\
                                & \checkmark       & \checkmark          & ~               & ~          & 92.23          & 85.59          & \textbf{92.70} & \textbf{86.39} & \textbf{97.74} & \textbf{95.57}  \\
                                & \checkmark        & ~          & \checkmark               & ~          & 92.07          & 85.29          & 90.10          & 81.99          & 97.42          & 94.91           \\
                                & \checkmark        & ~          & ~               & \checkmark          & \textbf{92.42} & \textbf{85.90} & 92.56          & 86.15          & 97.56          & 95.24           \\ 
\hline
\multirow{5}{*}{~
  ~
  ResNet} & ~        & ~          & ~               & ~          & 91.26          & 83.92          & 91.11          & 83.68          & 97.14          & 94.45           \\
                                & \checkmark        & ~          & ~               & ~          & 91.35          & 84.04          & 91.21          & 83.87          & 97.48          & 95.09           \\
                                & \checkmark        & \checkmark          & ~               & ~          & 91.85          & 84.92          & 91.23          & 83.86          & \textbf{98.11} & \textbf{96.28}  \\
                                & \checkmark        & ~          & \checkmark               & ~          & 91.62          & 84.54          & \textbf{92.06} & \textbf{85.28} & 97.36          & 94.87           \\
                                & \checkmark        & ~          & ~               & \checkmark          & \textbf{91.87} & \textbf{84.97} & 91.69          & 84.65          & 97.85          & 95.78           \\
\cline{1-11}
\end{tabular}
\end{table*}
Optimizing hard case samples in the later stages of model training is crucial for enhancing model performance. At this stage, background hard case samples should become the focus of model optimization. Therefore, we introduce dynamic weight functions to shift the learning focus of the model from the foreground to the background. We designed three types of dynamic weight functions and conducted experiments on three datasets. As shown in TABLE VII, in the experiments using ResNet as the backbone, the maximum F1 score improvement achieved by using dynamic weight functions compared to not using them was 0.52\%, 0.85\%, and 0.63\%, respectively; similar results were obtained in experiments using PVT as the backbone, with increases of 0.34\%, 0.80\%, and 0.54\%, respectively. This indicates that the method based on dynamic weighting reduces early training errors in predicting hard case samples, thereby enhancing the CD performance of the model; moreover, focusing on optimizing background hard case samples in the later stages of training is both reasonable and effective. According to both sets of experiments, although all three dynamic weight functions achieved some accuracy improvements, they had no absolute advantage. This might be due to the varying numbers and distributions of hard case samples in different datasets; hence, different dynamic weight functions are suitable for different datasets. We can choose the appropriate dynamic weight function based on data characteristics. For example, cosine dynamic weighting is more suitable for datasets with a clear gap between simple and hard case samples, allowing for stable adjustment of the loss distribution to achieve healthy convergence; exponential dynamic weighting is more suitable for situations where there are fewer foreground samples and more background samples.
\subsubsection{Effect of the focusing factor}
\begin{table*}
\centering
\caption{COMPARISON OF DIFFERENT FOCUSING FACTOR RESULTS}
\begin{tabular}{ccccccccc} 
\hline
Dataset                    & $\gamma$       & 0.0   & 0.3   & 0.5   & 1.0            & 2.0            & 4.0   & 6.0    \\ 
\hline
\multirow{2}{*}{LEVIR-CD}  & F1  & 92.23 & 92.25 & 92.27 & 92.30          & \textbf{92.42} & 92.27 & 92.17  \\
                           & IOU & 85.67 & 85.66 & 85.66 & 85.7           & \textbf{85.90} & 85.65 & 85.49  \\ 
\hline
\multirow{2}{*}{Google-CD} & F1  & 91.94 & 92.45 & 92.47 & \textbf{93.14} & 92.7           & 92.45 & 90.92  \\
                           & IOU & 85.09 & 85.97 & 85.99 & \textbf{87.17} & 86.39          & 85.95 & 83.36  \\
\hline
\end{tabular}
\end{table*}
Introducing the hard case awareness term $(1-p)^\gamma$ leverages the distribution of loss values to approximate the regions of hard case samples and then adjusts the attention weight distribution between hard case samples and foreground samples using the focus Factor $\gamma$. Generally, the larger $\gamma$ is, the greater the weight and the higher the degree of attention. We conducted comparative experiments with progressively increasing $\gamma$ on two datasets and observed similar experimental effects. As shown in TABLE VIII, starting from $\gamma$=0.0 and increasing $\gamma$, the CD accuracy continuously improved. On the LEVIR-CD and Google-CD datasets, when $\gamma$ was set to 2.0 and 1.0, respectively, the F1 scores were 0.19\% and 1.20\% higher than when $\gamma$ was 0.0, reaching absolute accuracies of 92.42\% and 93.14\%, respectively. However, as $\gamma$ continued to increase, the performance of the model began to decline. This may be due to two reasons: first, image noise was also given a high focus Factor $\gamma$, leading to its misclassification as hard case samples; second, an excessively large $\gamma$ caused the model to excessively neglect foreground targets, thus reducing its discriminative ability for foreground interest targets. From this, we can conclude that an appropriate focus Factor $\gamma$ can enhance the model's CD performance, but different data characteristics often require different optimal focus factors.

\subsection{Analysis of effectiveness}
\subsubsection{T-SNE and attention heatmap visualization effect}
To demonstrate how our method focuses on various remote sensing targets, we visualized the attention heatmap of the last layer of HSONet and overlaid it on the T2 image to obtain the image shown in (f). In (f), the attention distribution of the changing targets can be clearly observed, where red denotes higher attention values and blue denotes lower values. As (f) shows in Figures 13 and 14, areas with high heating values completely cover the areas of interest changes. The house change areas in LEVIR-CD and Google-CD both have high heating values, indicating that our method learned the feature representation of interest targets and is not sensitive to noninteresting changes. Notably, HSONet demonstrates strong robustness to building shadows, as shown in (2) and (3) of Figure 13. The shadows around buildings do not affect our method's ability to accurately detect real change areas. Moreover, shadows around buildings may be a key factor in change recognition because our model can implicitly learn certain additional feature concepts to facilitate the identification of change areas.
\begin{figure*}[ht]
    \centering
    \includegraphics[width=6in]{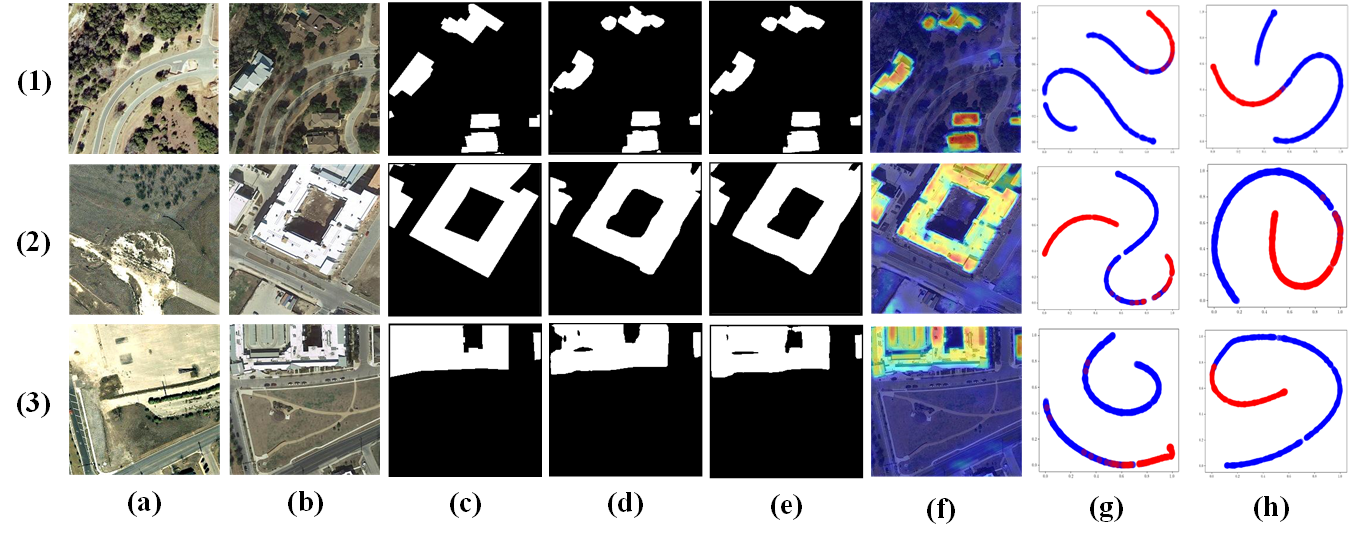}
    \caption{t-SNE visual comparison on LEVIR-CD:(a)T1, (b)T2, (c)GT, (d)ICIF-Net, (e)Our HSONet, (f)HSONet heatmap, (g)ICIF-Net t-SNE, and (h)HSONet t-SNE.}
    \label{fig13}
\end{figure*}
\begin{figure*}[ht]
    \centering
    \includegraphics[width=6in]{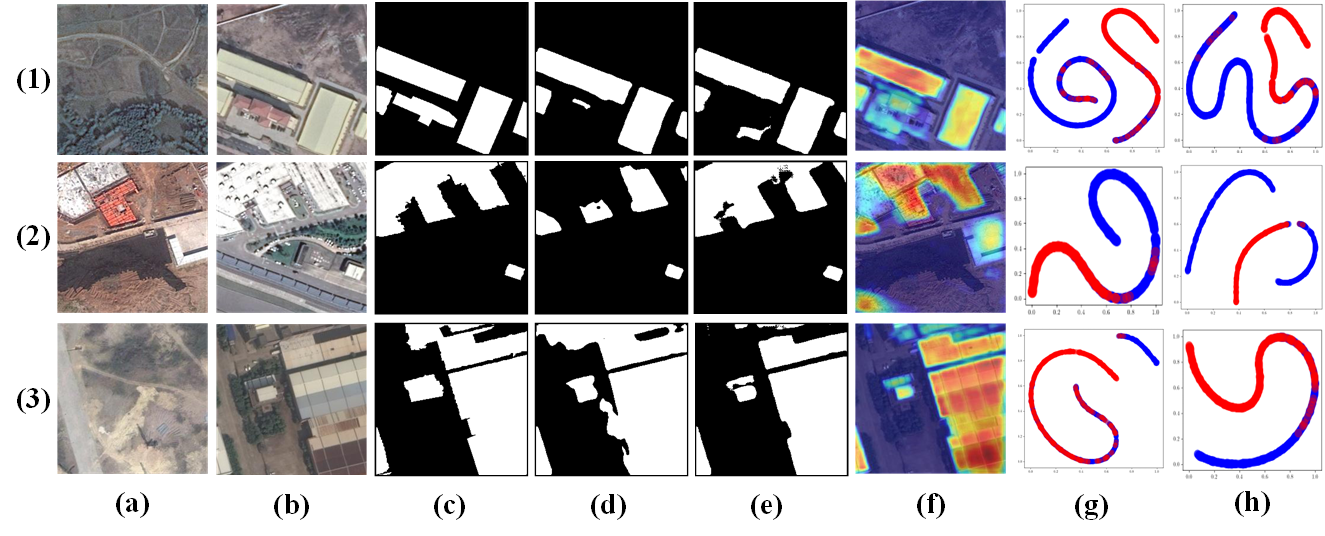}
    \caption{t-SNE visual comparison on Google-CD:(a) T1, (b)T2, (c)GT, (d)ICIF-Net, (e)Our HSONet, (f)HSONet heatmap, (g)ICIF-Net t-SNE, and (h)HSONet t-SNE.}
    \label{fig14}
\end{figure*}

In addition, to more intuitively reflect the degree of separation between the changed and unchanged samples, we used t-distributed stochastic neighbour embedding (t-SNE)\cite{50} to downscale the distribution of the visualized samples and compare and analyse our method with the distribution of the ICIF-Net features. As shown in (g) and (h) of Figures 14 and 15, red represents changed samples, and blue represents unchanged samples. Notably, in the dimensionality reduction results of HSONet, the separation boundary between the red and blue samples is clearer than that between the other samples, with both sample categories being more distinctly clustered and with few instances of red and blue intermingling. This indicates that our method has more accurate change perception and anti-interference capabilities, which is why HSONet's change confidence is greater and it has clear boundaries. In contrast, in the ICIF-Net dimensionality reduction results, a certain number of red samples appear in the blue samples, which indicates that some of the changed samples and the unchanged samples are indistinguishable in the feature space. This leads to some samples at the edge of the changed area with low confidence being misclassified by the network, resulting in "blurred boundaries of the changed area".
\subsubsection{Attention shift heatmap visualization}
To validate the attention shift pattern of the model during the early and late training stages, we visualized the attention focus distribution on sample data at different epochs. Each epoch is equidistant, allowing the observation of the pattern of shifting optimization focus. Figures 15, 16, and 17 show the heatmaps of a particular data sample in LEVIR-CD, Google-CD, and CDD, respectively, visualized using Grad-CAM\cite{51}. All three images exhibit the same pattern: in the early stages of training, the optimization focus is concentrated on interest targets in the foreground, which in this case are 'buildings'. By the middle stages of training, the model gradually shifts towards learning about background information, such as the areas around roads or less obvious buildings. In the later stages of training, the model's attention increasingly turns to learning about background features, during which there is a varying degree of neglect for foreground targets.
\begin{figure*}[ht]
    \centering
    \includegraphics[width=7in]{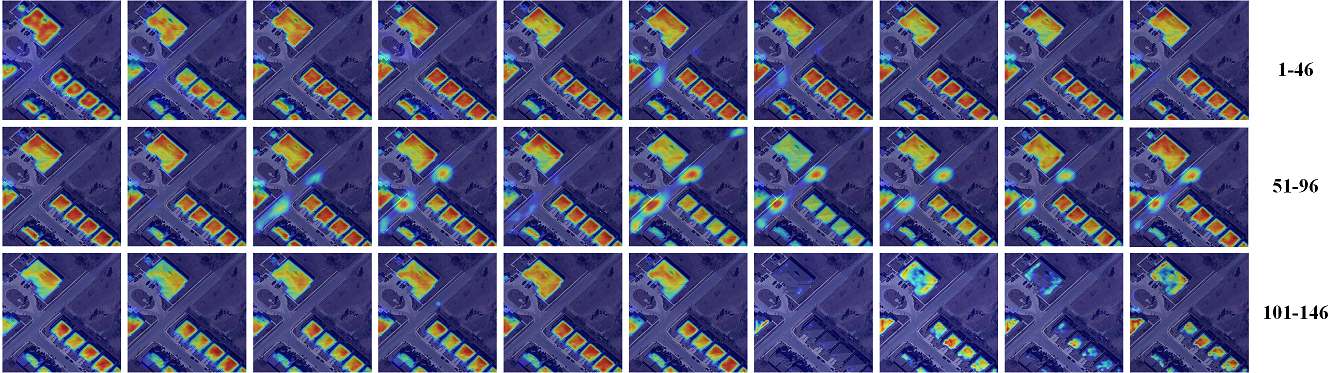}
    \caption{Attention shift heatmap on the LEVIR-CD validation set.}
    \label{fig15}
\end{figure*}
\begin{figure*}[ht]
    \centering
    \includegraphics[width=7in]{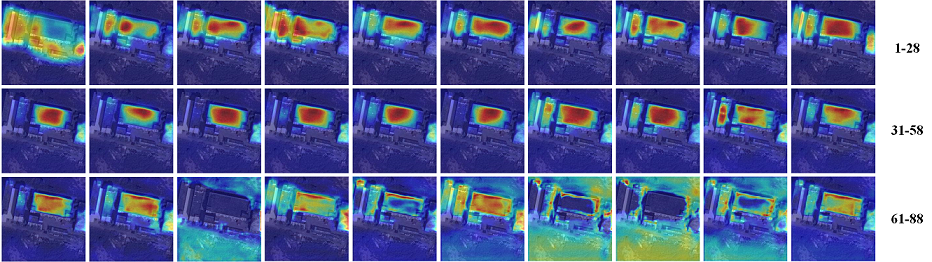}
    \caption{Attention shift heatmap on the Google-CD validation set.}
    \label{fig16}
\end{figure*}
\begin{figure*}[ht]
    \centering
    \includegraphics[width=7in]{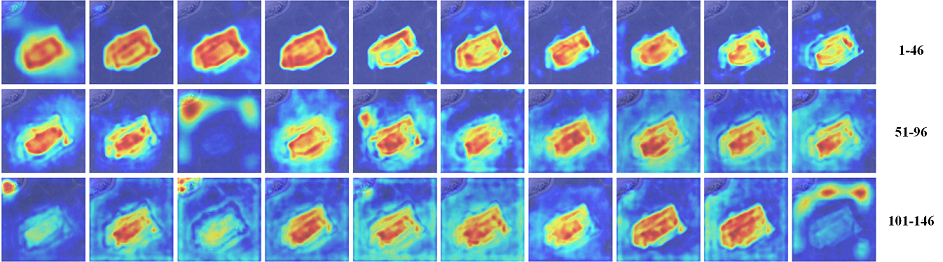}
    \caption{Attention shift heatmap on the CDD validation set.}
    \label{fig17}
\end{figure*}

\section{Conclusion and future work}
In response to the challenge of obtaining hard case samples from remote sensing imagery, this paper presents a Siamese foreground association-driven hard case sample optimization network, named HSONet, which includes a foreground-scene association module and an EO-loss. In the equilibrium optimization loss function, the distribution of hard case samples is estimated through the hard case awareness term, and the optimization focus of the network is shifted dynamically to balance the learning process for both foreground targets and background hard case samples. In the foreground-scene association module, the associations between interest targets in the foreground and related context in latent remote sensing spatial scene information are modelled, and background hard case samples are actively output to ensure the successful extraction of interest changes in both foreground and background hard case samples. This strategy effectively captures and perceives changing features, showing advantages in detecting changes in hard case samples. Experimental results on four change detection datasets show that our method obtains more accurate CD results than other methods, is less prone to problems such as omissions and misdetections, and is robust against hard case samples.

Additionally, our method is versatile. Although it is applied only to CD tasks here, the proposed strategy can be readily transferred to other application scenarios. Of course, supervised learning algorithms are limited by data availability. Therefore, in the future, we plan to explore optimization methods further for hard case samples in semisupervised\cite{qiji} or self-supervised learning\cite{caojun,zzy} contexts.

\bibliographystyle{unsrt}
\bibliography{cit}

\end{document}